\documentclass[jad,final]{iosart2x}
\usepackage{graphicx}

\usepackage{booktabs}
\usepackage{pdflscape}
\usepackage{threeparttable}
\usepackage{longtable}
\usepackage{tabularx}
\usepackage{soul}
\usepackage[textsize=tiny,
disable,
textwidth=2.1cm]{todonotes}
\usepackage{marginnote}

\graphicspath{{.}{figures/}}

\newcommand{\sfnote}[2][]{\hl{#1}\todo[line,color=green!10]{sf: #2}{}}

\makeatletter
\if@todonotes@disabled

\else

\fi
\makeatother

\makeatletter
\gdef\@ptsize{0}
\let\@currsize\normalsize
\makeatother
\usepackage{fancyhdr}

\newcommand{\citeAD}[1]{\citeauthor{#1}\setcitestyle{round} \citeyearpar{#1}\setcitestyle{square} \citep{#1}}

\pubyear{0000}
\volume{0}
\firstpage{1}
\lastpage{1}

\begin{document}
\makeatletter
\let\put@numberlines@box\relax
\makeatother
\begin{frontmatter}
\title{ Artificial Intelligence, speech and language
    processing approaches to monitoring Alzheimer's Disease: a
    systematic review}
\runtitle{AI approaches to monitoring AD}
\runauthor{de la Fuente Garcia, Ritchie \& Luz}


\author[A]{\inits{S.}\fnms{Sofia} \snm{\textbf{de la Fuente Garcia}}\ead[label=e1]{sofia.delafuente@ed.ac.uk}%
  \thanks{Corresponding author. \printead{e1}. Nine Edinburgh BioQuarter, 9 Little France Road, Edinburgh EH16 4UX}
},
\author[B]{\inits{C.}\fnms{Craig} \snm{\textbf{Ritchie}}\ead[label=e2]{craig.ritchie@ed.ac.uk}}
and
\author[A]{\inits{S.}\fnms{Saturnino} \snm{\textbf{Luz}}\ead[label=e3]{s.luz@ed.ac.uk}}
\address[A]{Usher Institute, Edinburgh Medical School, \orgname{The
    University of Edinburgh},
Scotland, \cny{UK}\printead[presep={\\}]{e1,e3}}
\address[B]{Centre for Dementia Prevention, \orgname{The University of
  Edinburgh},
Scotland, \cny{UK}\printead[presep={\\}]{e2}}

\begin{abstract}\\
\textbf{Background:} Language is a valuable source of clinical information in Alzheimer’s Disease, as it declines concurrently with neurodegeneration. Consequently, speech and language data have been extensively studied in connection with its diagnosis.  \\
\textbf{Objective:} firstly, to summarise the existing findings on
  the use of artificial intelligence, speech and language processing to predict cognitive
  decline in the context of Alzheimer's Disease. Secondly, to detail current research procedures, highlight their limitations and suggest strategies to address them.\\
  \textbf{Method:} Systematic review of original research between 2000 and 2019, registered in PROSPERO (reference CRD42018116606). An interdisciplinary search covered six
  databases on engineering (ACM and IEEE), psychology (PsycINFO),
  medicine (PubMed and Embase) and Web of Science. Bibliographies of relevant papers were screened until
  December 2019. \\
  \textbf{Results:} from 3,654 search results 51 articles were selected against
  the eligibility criteria. Four tables summarise their findings: \textit{study details}, (aim, population,
  interventions, comparisons, methods and outcomes), \textit{data details} (size, type, modalities, annotation, balance, availability and language of study),
   \textit{methodology} (pre-processing, feature generation, machine learning,
  evaluation and results) and \textit{clinical
    applicability} (research implications, clinical potential, risk of
  bias and strengths/limitations).\\
  \textbf{Conclusion:} promising results are reported
  across nearly all 51 studies, but very few have been implemented in
  clinical research or practice. The main limitations of the field
 are poor standardisation, limited comparability of results, and a degree of
  disconnect between study aims and clinical applications. Active
  attempts to close these gaps will support translation of future
  research into clinical practice.
\end{abstract}


\end{frontmatter}
\lhead{}\chead{}\rhead{}\renewcommand{\headrulewidth}{0pt}
\thispagestyle{fancy}



\section*{Introduction}

Alzheimer's Disease (AD) is a neurodegenerative disease that involves
decline of cognitive and functional abilities as the illness
progresses \cite{american2013diagnostic}. It is the most common
aetiology of dementia. Given its prevalence, it has effects beyond
just patients and carers as it also has a severe societal and economic
impact worldwide \cite{WorldHealthOrganization2013}. Although memory
loss is often considered the signature symptom of AD, language
impairment may also appear in its early stages
\cite{Ross1990}. Consequently, and due to the ubiquitous nature of
speech and language, multiple studies rely on these modalities as
sources of clinical information for AD, from foundational qualitative
research \cite[e.g.][]{watson1999analysis, bucks2000analysis} to more
recent work on computational speech technology
\cite[e.g.][]{luz2017longitudinal, Fraser2019,
  Mirheidari2019dementia}. The potential for using speech as a
biomarker for AD is based on several prospective values, including: 1)
the ease with which speech can be recorded and tracked over time, 2)
its non-invasiveness, 3) the fact that technologies for speech
analysis have improved markedly in the past decade, boosted by
advances in artificial intelligence (AI) and machine learning, and 4)
the fact that speech problems may be manifest at different stages of the disease, making it a life-course
assessment that has value unlimited by disease stage.

Recent studies on the use of AI in AD research entail using language
and speech data collected in different ways and applying computational
speech processing for diagnosis, prognosis or progression
modelling. This technology encompasses methods for recognizing,
analysing and understanding spoken discourse. It
implies that at least part of the AD detection process could be
automated (passive).  Machine learning methods have been central to
this research programme. Machine learning is a field of AI that
concerns itself with the induction of predictive models ``learnt''
directly from data, where the learner improves its
own performance through ``experience'' (i.e. exposure to greater
amounts of data).  Research on automatic processing of speech and
language with AI and machine learning methods have yielded
encouraging results and attracted increasing interest.
Different approaches have been studied, including computational
linguistics \cite[e.g.][]{Fraser2016}, computational paralinguistics
\cite[e.g.][]{LUZ18.5}, signal processing
\cite[e.g.][]{haider2019assessment} and human-robot interaction
\cite[e.g.][]{Mirheidari2019computational}. 

However, investigations of the use of language and speech technology
in Alzheimer's research are heterogeneous, which makes consensus,
conclusions and translation into larger studies or clinical practice
problematic. The range of goals pursued in such studies is also broad,
including automated screening for early Alzheimer's disease, tools for
early detection of disease in clinical practice, monitoring of disease
progression and signalling potential mechanistic underpinnings to
speech problems at a biological level thereby improving disease
models.  Despite progress in research, the small, inconsistent,
single-lab and non-standardised nature of most studies has yielded
results that are not robust enough to be aggregated and thereafter
implemented towards those goals. This has resulted in gaps between
research contexts, clinical potential and actual clinical applications
of this new technology.

We sought to summarise the current state of the evidence regarding AI
approaches in speech analysis for Alzheimer's disease with a view to
setting a foundation for future research in this area and potential
development of guidelines for research and implementation. The review
has three main objectives. Firstly, to present the main aims and
findings of this research, secondly to outline the main methodological
approaches and finally surmise the potential for each technique to be
ready for further evaluation towards clinical use. In doing so we hope
to contribute to the development of these novel, exciting, and yet
under-utilised approaches, towards clinical practice.

\section*{Methods}
The procedures adopted in this review were specified in a protocol
registered with the international prospective register of systematic
reviews
PROSPERO (reference: CRD42018116606).
In the following sections we describe the elegibility criteria,
information sources, search strategy, study records management, study
records selection, data collection process, data items (extraction
tool), risk of bias in individual studies, data synthesis,
meta-bias(es) and confidence in cumulative evidence.

\subsection*{Elegibility criteria}
We aimed to summarise all available scientific studies where an
interactive artificial intelligence (AI) approach was adopted for
neuropsychological monitoring. Interaction-based technology entails
data obtained through a form of communication, and AI entails some
automation of the process. Therefore, we included articles where
automatic machine learning methods were used for AD screening,
detection and prediction, by means of computational linguistics and/or
speech technology.

Articles were deemed eligible if they described studies of
neurodegeneration in the context of AD. That is, subjective cognitive impairment (SCI), mild cognitive impairment (MCI), AD or other
dementia-related terminology if indicated as AD-related in the full
text (e.g. if a paper title reads unspecified "dementia" but the
research field is AD). The included
  studies examined behavioural patterns that may precede overt
  cognitive decline as well as observable cognitive impairment in these
  neurodegenerative diseases. Related
conditions such as semantic dementia (a form of aphasia) or
Parkinson's Disease (a different neurodegenerative disease) formed part
of the exclusion criteria (except when in comorbidity with
AD). Language was not an exclusion criterion, and translation resources
were used as appropriate.

Another exclusion criterion is the exclusive use of traditional
statistics in the analysis. The inclusion criteria require at least
one component of AI, ML or big data, even if the study encompasses
traditional statistical analysis. Further exclusion criteria apply to
related studies relying exclusively on neuroimaging techniques such as
magnetic resonance imaging (MRI), with no relation to language or
speech, even if they do implement AI methods. The same applies to
biomarker studies (e.g. APOE genotyping). This review also excluded
purely epidemiological studies, that is, studies aimed at analysing the
distribution of the condition rather than assessing the potential of
AI tools for monitoring its progress.

In terms of publication status, we considered peer-reviewed journal
and conference articles only. Records that were not original research
papers were excluded (i.e. conference abstracts and systematic
reviews). In order to avoid redundancy, we assessed research by the
same group and excluded overlapping publications. This was assessed by
reading the text in full and selecting only the most relevant article
for review (i.e. most comprehensive and up to date). Due to limited
resources, we also excluded papers when full-texts were found
unavailable in all our alternative sources.

\begin{figure}[htb]
    \centering
    \includegraphics[width=1\linewidth]{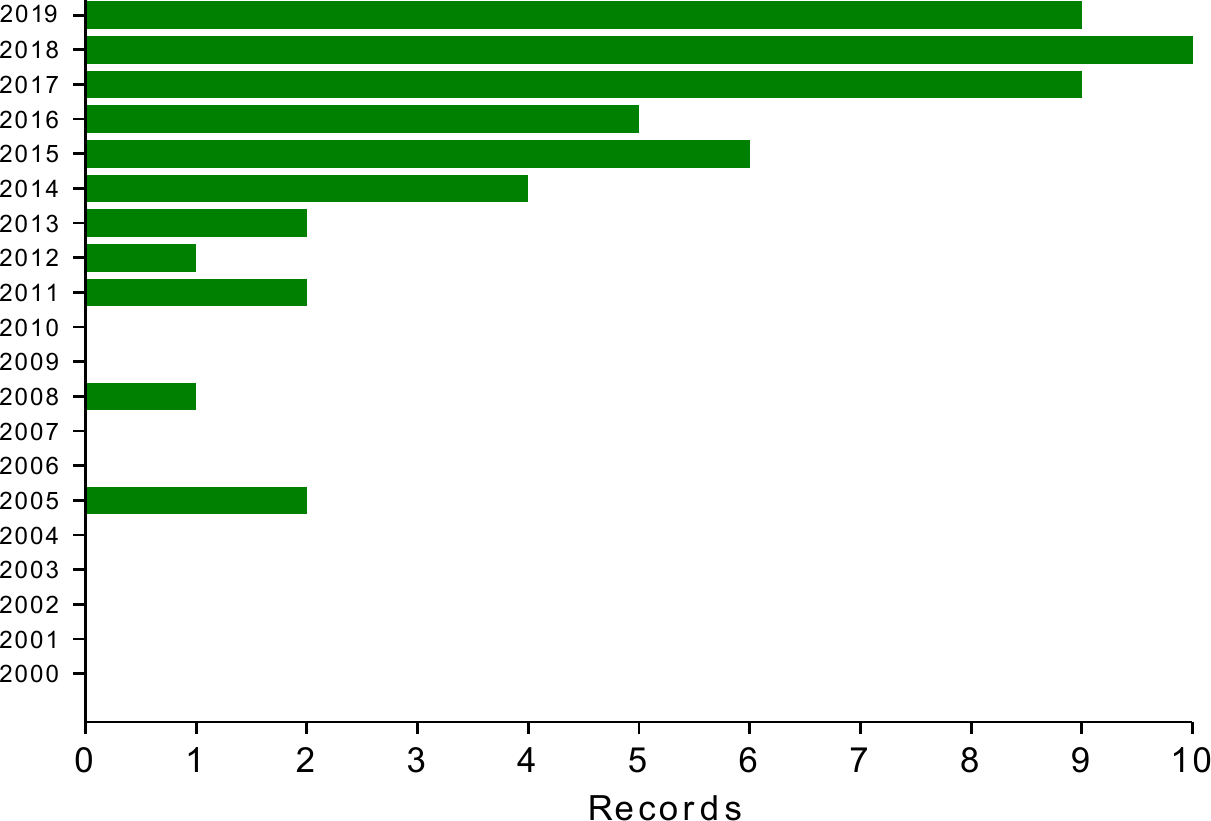}
    \caption{Number of records found suitable for review each year (2000-2019).}
    \label{fig:records}
\end{figure}

Lastly, we considered papers from a twenty-year span, from the
beginning of 2000 to the end of 2019, anticipating that the closer to
the end of this time-frame, the larger the number of
results, as shown in Figure~\ref{fig:records}.

\subsection*{Information Sources}
Between October and December 2019, we searched the following electronic
databases: ACM, Embase, IEEE, PsycINFO, PubMed, and Web of Science. We
contacted study authors by email when full-text versions of relevant
papers where not available through the university library, with
varying degrees of success.

We also included relevant titles found through "forward citation
tracking" with Google Scholar, screening articles references and
research portal suggestions suggestions.

\subsection*{Search Strategy}
Given the heterogeneity of the field, a broad search needed to be
conducted. For the health condition of interest, AD, we included terms
such as dementia, cognitive decline and Alzheimer. For the
methodology, we included speech, technology, analysis and natural
language processing, artificial intelligence, machine learning, and
big data.

The search strategy was developed collaboratively between the authors,
and with the help of the University of Edinburgh's academic support
librarian. After a few iterations and trials, we decided not to
include the AI terms, since this seemed to constrain the search too
much, yielding fewer results. Therefore, the search queries were
specified as follows (example for PubMed):
\begin{itemize}
\item (speech AND (dementia OR "cognitive decline" OR (cognit* AND
  impair*) OR Alzheimer) AND (technology OR analysis)) OR ("natural
  language processing" AND (dementia OR "cognitive decline" OR
  (cognit* AND impair*) OR Alzheimer) )
    \item Filters applied: 01/01/2000 - 31/12/2019.
\end{itemize}

Then, we applied the exclusion criteria, starting from the lack of AI,
ML and big data methods, usually detected in the abstract.

We used EndNote X8 \cite{analytics2016endnote} for study records management.

\subsection*{Study records selection}

Screening for record selection happened in two phases, independently
undertaken by two reviewers and following pre-established eligibility
criteria. In the first phase, the two independent authors screened
titles and abstracts against exclusion criteria using EndNote. The
second phase consisted of a full-text screening for those papers that
could not be absolutely included or excluded based on title and
abstract information only. Any emerging titles that were deemed
relevant were added to the screening process.  Disagreements at any of
the stages were discussed and, when necessary, a third author convened
to find a resolution. Some records reported results that were
redundant with a later paper of the same research group, mainly
because the earlier record was a conference paper or because an
extended version of the research paper had been published elsewhere at
a later date. When this happened, earlier and shorter reports were
excluded.

\subsection*{Data collection process}
Our original intention was to  rely on the PICO framework
\cite{richardson1995well} for data collection. However, given the
relative youth and heterogeneity of the research field reviewed, and the lack
of existing reviews on the topic, we adapted a data extraction tool
specifically for our purposes. This tool took the form of four comprehensive
tables which were used to extract the relevant information from each
paper. Those tables summarise general study information, data details,
methodology and clinical applicability.

The tables were initially ``piloted'' with a few studies, in order to
ensure they were fit to purpose. Information extraction was performed
independently by two reviewers and consistency was compared. When
differences about extracted items was not resolved by discussion, the
third author was available to mediate with the paper's full text as reference.

\subsection*{Data items (extraction tool)}
As stated in the data collection process, data items will be extracted
through the elaboration of four tables. These tables are:

\begin{itemize}
\item \textbf{SPICMO:} inspired in the PICO framework, it contains
  information on Study, Population, Interventions, Comparison groups,
  Methodology and Outcomes. More details can be found just before
  Table~\ref{tab:spicmo} (supplementary material).
    
\item \textbf{Data details:} dataset/subset size, data type, other
  data modalities, data annotation, data availability and
  language. More details can be found just before Table~\ref{tab:data} (supplementary material).
    
\item \textbf{Methodology details:} pre-processing, features
generated, ML task/method, evaluation technique and
  results. More details can be found just before Table~\ref{tab:methods} (supplementary material).

\item \textbf{Clinical applicability:} research implications, clinical
  potential, risk of bias, and strengths/limitations. More details can be
  found just before Table~\ref{tab:clinical} (supplementary material).

\end{itemize}

\subsection*{Risk of bias in individual studies}
Many issues, such as bias, do not apply straightforwardly to this review because it focuses on diagnostic and prognostic test accuracy, rather than interventions. Therefore, if there were to be significance tests they would be for comparisons between the results of the different methods. Besides, the scope of the review is
machine learning technology, where the evaluation through significance
testing is rare. Papers that rely exclusively on traditional
statistics will be excluded, and therefore we expect the
review to suffer from a negligible risk of bias in terms of
significance testing. 

The risk of bias in machine learning studies often comes from how the
data is prepared in order to train your models. In a brief example, if
a dataset is not split in a training and a testing subset, the model
will be trained and tested on the same data. Such model is likely to
achieve very good results, but chances are that its performance will
drop dramatically when tested on unseen data. This risk is called
"overfitting", and is assessed in the last table of the review
(Clinical Significance, Table~\ref{tab:clinical}). Other risks
accounted for in this table are data balance, the use of suitable
metrics, the contextualization of results and the sample size. Data
balance reports whether the dataset has comparable numbers of AD and
healthy participants, as well as in terms of gender or age. Suitable
metrics is an assessment of whether the metric chosen to evaluate a
model is appropriate, in conjunction with data balance and
sample size (e.g. accuracy is not a robust metric when a dataset is
imbalanced). Contextualization refers to whether their study results
are compared to a suitable baseline (i.e. a measure without a target
variable or comparable research results). Finally, sample size is
particularly relevant because machine learning methodology was
developed for large datasets, but data scarcity is a distinctive
feature of this field.

The poor reporting of results and subsequent interpretation
difficulties is a longstanding challenge of diagnostic test accuracy
research \cite{leeflang2008systematic}. Initially, we considered two
tools for risk of bias assessment, namely the "QUADAS-2: Quality
Assessment of Diagnosis Studies checklist - 2"
\cite{whiting2011quadas} and the "PROBAST: Prediction model Risk Of
Bias ASsessment Tool" \cite{wolff2019probast}. However, our search
covers an emerging interdisciplinary field where papers are neither
diagnostic studies nor predictive ones. Additionally, the Cochrane
Collaboration recently emphasised a preference for systematic reviews
to focus on the performance of individual papers' on the different
risk of bias criteria \cite{higgins2008assessing}. Consequently, we
decided to assess risk of bias as part of the Clinical Applicability
table (table~\ref{tab:clinical}), according to criteria that are
suitable to the heterogeneity currently inherent to the field. These
criteria include the risks of bias described above, as well as an
assessment of generalisability, replicability and validity, which are
standard indicators of the quality of a study. Risk of bias was
independently assessed by two reviewers and disagreements were
resolved by discussion.

\subsection*{Data synthesis}
Given the discussed characteristics of the field, as well as the broad
range of details covered by the tables, we anticipate a thorough
discussion of all the deficiencies and inconsistencies that future
research should address. Therefore, we summarise the data in narrative
form, following the structure provided by the features summarised in
each table. Although a meta-analysis is beyond scope at the current
stage of the field, we do report outcome measures in a comparative
manner when possible.

\subsection*{Confidence in cumulative evidence}
We will assess accuracy of prognostic and diagnostic tools, rather
than confidence in an intervention. Hence, we will not be drawing any
conclusions related to treatment implementation.

\subsection*{Background on AI, Cognitive tests and Databases} \label{sec:background}

This section briefly defines key terminology and abbreviations
referring and offers a taxonomy of features, adapted from
\citet{voleti2019review}, to enhance the readability of the systematic
review tables.  This section also briefly
describes the most commonly used databases and neuropsychological
assessments, with the intention of making these accessible for the
reader.

\subsubsection*{AI, machine learning, and speech technologies}\label{subsec:AI}

AI can be loosely defined as a field of research that studies
artificial computational systems that are capable of exhibiting
human-like abilities or human level performance in complex
tasks. While the field encompasses a variety of symbol manipulation
systems and manual encoding of expert knowledge, the majority of
methods and techniques employed by the studies reviewed here concern
machine learning methods. While machine learning dates back to the
50's, the term ``machine learning'' as it is used today, originated
within the AI community in the late 70's to designate a number of
techniques designed to automate the process of knowledge
acquisition. Theoretical developments in computational learning theory
and the resurgence of connectionism in the 80's helped consolidate the
field, which incorporated elements of signal processing, information
theory, statistics and probabilistic inference, as well as inspiration
from a number of disciplines.

The general architecture of a machine learning system, as used in AD
prediction based on speech and language can be described in terms of the
learning task, data representation, learning algorithm, nature of
the ``training data'' and performance measures. The learning task
concerns the specification of the function to be learnt by the
system. In this review, such functions include classification (for
instance, the mapping of a voice or textual sample from a patient to a
target category such as ``probable AD'', ``MCI'' or ``healthy
control'') and regression tasks (such as mapping the same kind of
input to a numerical score, such as a neuropsychological test
score). The data representation defines which features of the vocal or
linguistic input will be used in the mapping of that input to the
target category or value, and how these features will be formally
encoded. Much research in machine learning applied to this and other
areas focuses on data representation. A taxonomy of features used in
the papers reviewed here is presented on
table~\ref{tab:FeatureTaxonomy}. There is a large variety of learning
algorithms available to the practitioner, and a number of them have
been employed in AD research. These range from connectionist systems,
of which most ``deep learning'' architectures are examples, to
relatively simple linear classifiers such as na\"{i}ve Bayes and
logistic regression, to algorithms that produce interpretable outputs
in the form of decision trees or logical expressions, to ensembles of
classifiers and boosting methods. The nature of the training data
affects both its representation and the choice of algorithm. Usually,
in AD research, patient data are annotated with labels for the target
category (e.g. ``AD'', ``control'') or numerical scores. Machine
learning algorithms that make use of such annotated data for induction
of models are said to perform supervised learning, while learning that
seeks to structure unannotated data is called unsupervised
learning. Performance measures, and by extension the loss function
with respect to which the learning algorithm attempts to optimise,
usually depend on the application. Commonly used performance measures
are accuracy, sensitivity (also known as recall), specificity,
positive predictive value (also known as precision), and summary
measures of trade-offs between these measures, such as area under the
receiver operating characteristic curve, and F scores. These methods
and metrics are further detailed below.   

\setlength{\footskip}{3.5cm}
\begin{table*}[tbh!]
  \caption{Feature taxonomy, adapted from \citet{voleti2019review}.}\label{tab:FeatureTaxonomy}
  \raggedleft
\footnotesize
\begin{tabular}{ p{1.5cm}p{2.6cm}p{4.5cm}p{5cm}  }
  \hline
 Category   & Subcategory &  Feature type & Feature name, abbreviation, reference.\\
 \hline
 Text-based & Lexical features & Bag of words, vocabulary analysis & \textit{BoW}, \textit{Vocab}.\\
    (NLP)   &                 & Linguistic Inquiry and Word Count & \textit{LIWC} \cite{tausczik2010psychological}\\
            &                 & Lexical diversity                 & Type-Token Ratio (\textit{TTR}),\newline 
                                                                    Moving Average TTR (\textit{MATTR}),\newline 
                                                                    Simpson's Diversity Index (\textit{SDI})\newline
                                                                    Brunét's Index (\textit{BI}),\newline 
                                                                    Honoré's Statistic (\textit{HS}).\\
            &                 & Lexical Density                   & Content density (\textit{CD}),\newline 
                                                                    Idea Density (\textit{ID}),\newline 
                                                                    \textit{P}-Density (\textit{PD}).\\
            &                 & Part-of-Speech tagging &\textit{PoS}.\\
  \cline{2-4}
  
            & Syntactical features & Constituency-based parse tree scores & \textit{Yngve} \cite{yngve1960model},\newline
                                                                            \textit{Frazier} \cite{frazier1985}. \\
            &                      & Dependency-based parse tree scores   & \\
            &                      & Speech graph                         & Speech Graph Attributes (\textit{SGA}).\\
  \cline{2-4}
            & Semantic features  &  Matrix decomposition methods  & Latent Semantic Analysis (\textit{LSA}),\newline Principal Component Analysys (PCA).\\
            & (Word and sentence   & Neural word/sentence embeddings & \textit{word2vec}                   \cite{mikolov2013efficient}\\ 
  
            & embeddings)          & Topic modelling & \textit{Latent
                                                       Dirichlet Allocation} \cite{blei2003latent}.\\
            &                                         & Psycholinguistics & Reliance on familiar words (\textit{PsyLing}). \\
    \cline{2-4}
            & Pragmatics & Sentiment analysis & \textit{Sent}.\\
            &             & Use of language \textit{UoL} & Pronouns, paraphrasing, filler words (\textit{FW)}.\\
            & & Coherence & \textit{Coh}.\\
  
  \hline
  Acoustic & Prosodic features & Temporal &  Pause rate (\textit{PR}), \newline
                                            Phonation rate (\textit{PhR}), \newline
                                            Speech rate (\textit{SR}),\newline
                                            Articulation rate (\textit{AR}).\newline
                                            Vocalisation events.\\
            &      & Fundamental Frequency &  $F_0$ and trajectory. \\
            &      & Loudness and energy   &  \textit{loud}, \textit{E}.\\
            &      & Emotional content     &  \textit{emo}.\newline\\
  \cline{2-4}
            & Spectral features & Formant trajectories &  $F_1$, $F_2$, $F_3$.\\
            &                   & Mel Frequency Cepstral Coefficients  &  \textit{MFCCs} \cite{davis1980comparison}.\\
  
  \cline{2-4}
            & Vocal quality     & Jitter, Shimmer, harmonic-to-noise ratio &  \textit{jitt}, \textit{shimm}, \textit{HNR}.\\
  \cline{2-4}
            & ASR-related       & Filled pauses, repetitions, dysfluencies, hesitations. fractal dimension, entropy.                            & \textit{FP, rep, dys, hes, FD, entr}.\\
            &                   & Dialogue features (i.e. Turn-Taking)     & \textit{TT}:avg turn length, inter-turn silences.\\
  \hline
\end{tabular}
\end{table*}

\subsubsection*{Cognitive tests}\label{subsec:cognitivetests}
This is a brief description of the traditional cognitive tests (as opposed to speech-based cognitive tasks) most commonly
applied in this field, with two main purposes. On the one hand, neuropsychological
assessments are one of the several factors on which clinicians rely in order to make a clinical diagnosis, which in turn results on participants being assigned to an experimental group  (i.e. healthy control, SCI,
MCI, or AD). On the other hand, some of these tests are recurrently used as part of the
speech elicitation protocols.

Batteries used for diagnostic purposes consist of reliable and
systematically validated assessment tools that evaluate a range of
cognitive abilities. They are specifically designed for dementia and
aimed to be time-efficient, as well as able to highlight preserved and
impaired abilities.  The most commonly used batteries are the Mini-Mental State
Examination \cite[MMSE;][]{folstein1975mini}, the Montreal Cognitive Assessment
\cite[MoCA;][]{nasreddine2005montreal}, the Hierarchical Dementia
Scale-Revised \cite[HDS-R;][]{cole2015hierarchic}, the Clinical
Dementia Rating \cite[CDR;][]{morris1991clinical}, the Clock Drawing
Test \cite[CDT;][]{freedman1994clock}, the Alzheimer's Disease
Assessment Scale, Cognitive part \cite[ADAS-Cog][]{rosen1984new}, the
Protocol for an Optimal Neurpsychological Evaluation \cite[PENO, in
French;]{joanette1995evaluation} or the General Practitioner
Assessment of Cognition \cite[GPCog;][]{brodaty2002gpcog}. Most of these tests have been translated into different languages, such as the Spanish version of the MMSE
\cite[MEC;][]{pena2012evaluacion}, which is used in a few reviewed papers.

Tools measuring general functioning, such as the General Deterioration
Scale \cite[GDS; ][]{reisberg1988global} or Activities of Daily Living,
such as the Katz Index \cite{wallace2007katz} and the Lawton Scale
\cite{graf2009lawton}, are also commonly used. Based on the results of
these tests, clinicians usually proceed to diagnose MCI, following
Petersen's criteria \cite{petersen1997aging}, or AD, following NINCDS-ADRDA
criteria \cite{mckhann1984clinical}. Alternative diagnoses appear
in some texts, such as Functional Memory Disorder (FMD), following
\citet{schmidtke2008syndrome}'s criteria.

Speech elicitation protocols often include tasks extracted from
examinations that were originally designed for aphasia, such as
fluency tasks. Semantic verbal fluency tasks \cite[SVF, in
COWAT;][]{benton1976multilingual} and are often known as ``animal
naming'' because they require the participant generating a list of
nouns from a certain category (e.g. animals) while being
recorded. Another tool recycled from aphasia examinations is the
Cookie Theft Picture task \cite{goodglass1983boston}, which requires
participants to describe a picture depicting a dynamic scene, and
hence to also elaborate a short story. Although that is by far the
most common picture used in such tests, other pictures have also been
designed to elicit speech in a similar way
\cite[e.g.][]{Sadeghian2017}.

Another group of tests consists, essentially, of language sub-tests
(i.e. vocabulary) and immediate/delayed recall tests, extracted from
batteries to measure intelligence and cognitive abilities, such as
the Wechsler Adult Intelligence Scale \cite[WAIS-III;][]{wechsler1997wais}
or the Wechsler Memory Scale \cite[WMS-III;][]{wechsler1997wms},
respectively. Besides, the National Adult Reading Test
\cite[NART;][]{nelson1991national}, the Arizona Battery for
Communication Disorders of Dementia ABCD battery
\cite[ABCD;][]{bayles1993abcd}, the Grandfather Passage,
\cite{darley1975motor} and a passage of The Little Prince
\cite{Mirzaei2018} are also used to elicit speech in some articles.

\subsubsection*{Databases}
\label{subsec:databases}
Although types of data will be further discussed later, we
hereby give an overview of the main datasets described. For space
reasons, we only mention here those datasets which have been used in
more than one study, and for which a requesting procedure might be
available. For monologue data:
\begin{itemize}
\item \textit{Pitt Corpus:} by far the most commonly used. It consists
  of picture descriptions elicited by the Cookie Theft Picture,
  generated by healthy participants and patients with probable AD, and
  linked to their neuropsychological data (i.e. MMSE). It was
  collected by the University of Pittsburgh \cite{Becker1994} and
  distributed through
  DementiaBank \cite{bib:MacWhinney19unt}.
    
\item \textit{BEA Hungarian Dataset:} this is a phonetic database,
  containing over 250 hours of multipurpose Hungarian spontaneous
  speech. It was collected by the Research Institute for Linguistics
  at the Hungarian Academy of Sciences \cite{gosy2013bea} and
  distributed through
  META-SHARE.
    
\item \textit{Gothenburgh MCI database:} this includes comprehensive
  assessments of young elderly participants during their Memory Clinic
  appointments and senior citizens that were recruited as their healthy
  counterparts \cite{wallin2016gothenburg}. Speech research undertaken
  with this dataset uses the Cookie Theft picture description and
  reading tasks subsets, all recorded in Swedish.
\end{itemize}

For dialogue data, the \textit{Carolina Conversations Collection (CCC)} is
the only available database. It consists of conversations 
between healthcare professionals and patients suffering from a chronic
disease, including AD. For dementia research, participants are
assigned to an AD group or a non-AD group, if their chronic condition
is unrelated to dementia (i.e. diabetes, heart disease). Conversations
are prompted by questions about their health condition and experience
in healthcare. It is collected and distributed by the Medical
University of South Carolina \cite{pope2011finding}.

In addition, some of the reviewed articles refer to the \textit{IVA
  dataset}, which consists of structured interviews undertaken and
recorded simultaneously by an Intelligent Virtual Agent (a computer
``avatar'') \cite{mirheidari2017avatar}. However, the potential
availability of this dataset is unknown.

\section*{Results}
\begin{figure*}[h!tb]
    \centering
    \includegraphics[width=\linewidth]{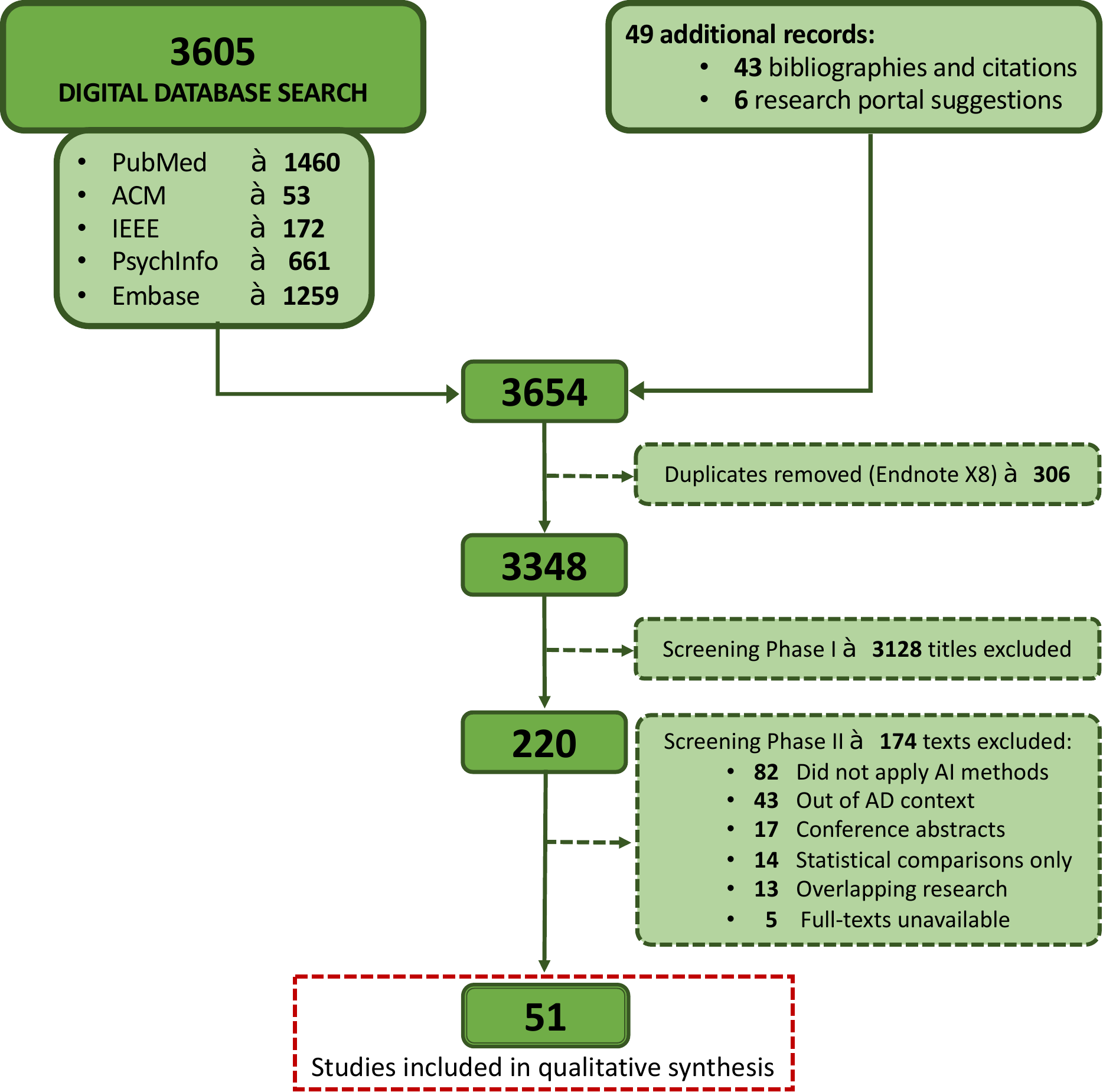}
    \caption{Screening and selection procedure, following guidelines provided by PRISMA \cite{moher2009preferred}.}
    \label{fig:flowchart}
\end{figure*}

Adding up all digital databases, the searches resulted in 3,605
records. Another 43 papers were identified by searching through bibliographies
and citations and 6 through research portal suggestions, adding up to
3,654 papers in total. Of those, 306 duplicates were removed using EndNote X8,
leaving 3,348 for the first screening phase. In this first phase, 3,128
papers were excluded based on title and abstract, and therefore 220
reached the second phased of screening. Five of these papers did not
have a full-text available, and therefore 215 papers where fully
screened. Finally, 51 papers were included in the review (Figure
\ref{fig:flowchart}).

\subsection*{Existing literature}

The review by \citet{voleti2019review} is to our knowledge, the only
published work with a comparable aim to the the present review,
although there are important scope differences. First of all, the
review by Voleti et Al. differs from ours in terms of methodological
scopes. Whilst their focus was to create a taxonomy for speech and
language features, ours was to survey diagnosis and cognitive
assessment methods that are used in this field and to assess the extent
to which they are successful. In this sense, our search was intentionally
broad. There are also differences in the scope of medical
applications. Their review studies a much broader range of disorders,
from schizophrenia to depression and cognitive decline. Our search,
however, targeted cognitive decline in the context of dementia and
Alzheimer's Disease. It is our belief that these reviews complement
each other in providing systematic accounts of these emerging fields.

\subsection*{Data extraction}
\label{sec:extraction}
Tables with information extracted from the papers are available as supplementary material. There are 4 different tables: a general table concerning usual
clinical features of interest (after the PICOS framework), and three
more specific tables concerning data details, methodology details and
implications for clinicians and researchers. Certain conventions
and acronyms were adopted when extracting article information, and should be considered when interpreting
the information contained on those tables. These conventions are available in the supplementary material, prior to the tables.



\section*{Discussion}

In this section, the data and outcomes of the different tables are synthesized in
different subsections and put into perspective. Consistent patterns
and exceptions are outlined. Descriptive aspects are organised by
column names, following table order and referencing their
corresponding table in brackets.

\subsection*{Study aim and design (table~\ref{tab:spicmo}: SPICMO)}

Most of the reviewed articles aim to use acoustic and/or linguistic
features in order to distinguish the speech produced by healthy
participants from the one produced by participants with a certain
degree of cognitive impairment. The majority of studies attempt binary
models to detecting AD and, less often, MCI, in comparison to HC. A
few studies also attempt to distinguish between MCI an AD. Even when
the dataset contains three or four groups (e.g. HC, SCI, MCI, AD),
most studies only report pairwise group comparisons \cite{Bertola2014,
  Konig2018, LundholmFors2018, MartinezdeLizarduy2017, Satt2013}. Out
of 51 reviewed papers, only seven did attempt three-way
\cite{Mirzaei2018, EgasLopez2019, Gosztolya2019, Kato2013} or four-way
\cite{Nasrolahzadeh2018, Mirheidari2019computational, Thomas2005}
classification. Their results are inconclusive and present potential
biases related to the quality of the datasets (i.e. low accuracy on
balanced datasets, or high accuracy on imbalanced datasets).

Slightly different objectives are described by \citet{Clark2016}, the
only study predicting conversion from MCI to AD, and by
\citet{Weiner2016}, the only study predicting progression from HC to
any form of cognitive impairment. While these studies also learnt
classifiers to detect differences between groups, they differ from
other studies in that they use longitudinal data. There is only one
article with a different aim than classification. This is the study by
\citet{duong2005heterogeneity}, who attempt to describe AD and HC
discourse patterns through cluster analysis.

Despite many titles mentioning cognitive monitoring, most research
addresses only the presence or absence of cognitive impairments (41,
out of 51 papers). Outside of those, seven papers are concerned with
three or four disease stages \cite{Mirzaei2018, EgasLopez2019,
  Gosztolya2019, Kato2013, Nasrolahzadeh2018,
  Mirheidari2019computational, Thomas2005}, two explore longitudinal
cognitive changes (although still through binary classification)
\cite{Clark2016,Weiner2016} and one describes discourse patterns
\cite{duong2005heterogeneity}. We note that future research could
take further advantage of this longitudinal aspect to build models
able to generate a score reflecting risk of developing an impairment.

\subsection*{Population (table~\ref{tab:spicmo}: SPICMO)}

The target population are elderly people who are healthy or exhibit
certain signs of cognitive decline related to AD (i.e. SCI, MCI,
AD). Demographic information is frequently reported, most commonly
age, followed by gender and years of education.

Cognitive scores such as MMSE are often part of the descriptive
information provided for study participants as well. This serves group
assignment purposes and allows quantitative comparisons of
participants' degree of cognitive decline. In certain studies, MMSE is
used to calculate the baseline against which classifier performance
will be measured \cite{Prudhommeaux2015,Sadeghian2017,Shinkawa2019}. However, despite being widely used in clinical and
epidemiological investigations, MMSE has been criticised for having
ceiling effects, especially when used to assess pre-clinical AD
\cite{carnero2014should}.

Some studies report no demographics \cite{luz2017longitudinal,
  PrudHommeaux2011, Thomas2005, Weiner2016, Weiner2016a}, only age
\cite{LUZ18.5, orimaye2017predicting}, only age and gender
\cite{Rochford2012, Troger2017, Satt2013}, or only age and education
\cite{Roark2011, Prudhommeaux2015}. An exception is the dataset
AZTIAHORE \cite{Lopez-de-Ipina2015a, Lopez-de-Ipina2015b}, which
contains the youngest healthy group (20-90 years old) and a typical AD
group (68-98 years old), introducing potential biases due to this
imbalance. Demographic variables are established risk factors for AD
\cite{mielke2014clinical}, therefore demographics reporting is
essential for this type of study.

\subsection*{Interventions (table~\ref{tab:spicmo}: SPICMO)}

Study interventions almost invariably consist of a speech generation
task preceded by a health assessment.  This varies between general
clinical assessments, including medical and neurological
examinations, and specific cognitive testing.  The comparison groups
are based on diagnosis groups, which in turn are established with the
results of such assessments. Therefore, papers lacking that
information do not specify their criteria for group assignment
\cite{Chien2018, EspinozaCuadros2014, Khodabakhsh2015,
  Lopez-de-Ipina2015a, Lopez-de-Ipina2015b, MartinezdeLizarduy2017,
  mirheidari2018detecting, orimaye2017predicting, Satt2013}. This
could be problematic, since the field currently revolves around
diagnostic categories, trying to identify such categories through speech
data. Consequently, one should ensure that standard criteria have been
used and that models are accurately tuned to these categories.

Speech tasks are sometimes part of the health assessment. For
instance, speech data are often recorded during the language sub-test
of a neuropsychological battery (e.g. verbal fluency, story recall or
picture description tasks). Another example of speech generated within
clinical assessment is the recording of patient-doctor consultations
\cite{mirheidari2017toward, mirheidari2018detecting,
  Mirheidari2019dementia} of cognitive examinations
\cite[e.g. MMSE,][]{EspinozaCuadros2014}. There are also studies where
participants are required to perform language tests outwith the health
assessment, for speech elicitation purposes only. Exceptionally, two
of these studies work with written rather than spoken language
\cite{fraser2019multilingual, Rentoumi2017}. Alternative tasks for
this purpose are reading text passages aloud
\cite[e.g.][]{Gonzalez-Moreira2015}, recalling short films
\cite[e.g.][]{Gosztolya2019}, retelling a story
\cite[e.g.][]{DosSantos2017}, retelling a day or a dream
\cite[e.g.][]{Beltrami2016}, or taking part in a semi-standardised
\cite[e.g.][]{Weiner2016} or conversational \cite[e.g.][]{LUZ18.5}
interview.

Most of these are examples of constrained, laboratory-based
interventions, which seldom include spontaneously generated
language. There are advantages to collecting speech under these
conditions, such as ease of standardisation, better control over
potential confunding factors, and focus on high cognitive load tasks
that may be more likely to elicit cognitive deficits. However,
analysis of spontaneous
speech production and natural conversations also has advantages. 
Spontaneous and conversational data can be captured in natural
settings over time, thus mitigating problems that might
affect performance in 
controlled, cross-sectional data, such as a participant having an
``off day'' or having slept poorly the night before the test.


\subsection*{Comparison groups (table~\ref{tab:spicmo}: SPICMO)}

This review targets cognitive decline in the context of AD. For its
purpose, nomenclature heterogeneity has been homogenised into four
consistent groups: HC, SCI, MCI and AD; with an additional group, CI,
to account for unspecified impairment (see
Table~\ref{tab:dx+terms}). As an exception to this nomenclature are
\citet{mirheidari2017toward,Mirheidari2019dementia,Mirheidari2019computational},
who compare participants with an impairment caused by
neurodegenerative disease (ND group, including AD) to an impairment
caused by functional memory disoders (FMD); and \citet{Weiner2016,
  Weiner2016a}, who introduce a category called age-associated
cognitive decline (AACD).

Furthermore, some studies add subdivisions to these categories. For
instance, there are two studies that classify different stages within
the AD group \cite{Lopez-de-Ipina2015a, Lopez-de-Ipina2015b}. Another
study divides the MCI group between amnesic single domain (aMCI) and
amnesic multiple domain (a+mdMCI), although classification results for
two groups are not very promising \cite{Bertola2014}. Within-subject
comparisons have also been attempted, comparing participants who
remained in a certain cognitive status to those who changed
\cite{Clark2016, Weiner2016a}.

Most studies target populations where a cohort has already been
diagnosed with AD or a related condition, looking for speech
differences between those and healthy cohorts. Therefore, little
insight is offered into pre-clinical stages of the disease.

\subsection*{Outcomes of interest (table~\ref{tab:spicmo}: SPICMO)}
    
Given the variety of diagnostic categories and types of data and
features used, it is not easy to establish state-of-the-art
performance. For binary classification, the most commonly attempted
task, the reported performance ranges widely depending in the data
use, the recording conditions, and the variables used in
modelling. For instance \citet{Lopez-de-Ipina2015b} reported an
accuracy that varied between 60\% and 93.79\% using only acoustic
features that were generated \textit{ad hoc}. Although the second
figure is very promising, their dataset is small, 40 participants, and
remarkably imbalanced in terms of both diagnostic class and age. In
terms of class, even though they initially report 20 AD and 20 HC, the
AD group is divided in three different severity stages, with 4, 10 and
6 participants respectively, whereas the control group remains
unchanged (20). In terms of age, 25\% percent of their healthy
controls fall within a 20-60 years old age range, while 100\% of the
AD group are over 60 years old. In contrast,
\citet{haider2019assessment} reported 78.7\% accuracy, using also
acoustic features only, but generated from standard feature sets that
had been developed for computational paralinguistics. Besides, this
figure appears as more robust because the dataset is much larger (164
participants) and it is balanced for class, age and gender, as well as
audio enhanced.  \citet{Guo2019} obtained 85.4\% accuracy on the same
dataset as \cite{haider2019assessment}, but using text-based features
only and without establishing class, age or gender balance. All the
figures quoted so far refer to monologue studies. The state-of-the-art
accuracy for dialogue data is 86.6\%, obtained by \citet{LUZ18.5}
using acoustic features only.

Regarding other classification experiments, we see that
\citet{Mirzaei2018} reports 62\% for a 3-way classification,
discriminating HC, MCI, AD. They are also among the few to
appropriately report accuracy, since they work with a class-balanced
dataset, while many other studies report overall accuracy in
class-imbalanced datasets. Accuracy figures can be very misleading in
the presence of class imbalance. A trivial rejector (i.e. a classifier
that trivially classifiers all instances as negative with respect to a
class of interest), would achieve very high accuracy on a dataset that
contained, say, 90\% negative instances. For example,
\citet{Nasrolahzadeh2018} report really high accuracy with a 4-way
classifier, 97.71\%, but in a highly imbalanced dataset. However,
\citet{Mirheidari2019computational} reported 62\% accuracy and 0.815
AUC for a 4-way classifier in a slightly more balanced
dataset. \citet{Thomas2005} also 4-way, only 50\%, on four groups of
MMSE scores.  Other studies attempting 3-way classification
experiments in balanced datasets are \citet{EgasLopez2019}, 56\% and
\citet{Gosztolya2019} with 66.7\%. \citet{Kato2013}, however, reports
85.4\% 3-way accuracy in an imbalanced dataset.

These results are diverse, and it stands clear that some will lead to
more robust conclusions than others.  Notwithstanding, numerical
outcomes are always subject to the science behind them, the quality of
the datasets and the rigour of the method. This disparity of results
therefore highlights the need for improved standards of reporting in
this kind of study. Reported results should include metrics that allow
the reader to assess the trade-off between false positives and false
negatives in classification, such as specificity, sensitivity, fallout
and F scores, as well measures that are less sensitive to class
imbalance, widely used in other applications of computational
paralinguistics, such as unweighted average recall. Contingency tables
and ROC curves should also be provided whenever possible. Given the
difficulties in reporting, comparing and differentiating the results
for the 51 reviewed studies on an equal footing, we refer the reader
to Tables~\ref{tab:spicmo} and \ref{tab:methods}, available in the
suplementary material for details.


\subsection*{Size of dataset or subset (table~\ref{tab:data}: Data Details)}

Within a machine learning context, all the reviewed studies use
relatively small datasets. About 31\% train their models with less
than 50 participants \cite{Beltrami2016, Chien2018, DArcy2008,
  EspinozaCuadros2014, Gonzalez-Moreira2015, Kato2013,
  Lopez-de-Ipina2015a, Lopez-de-Ipina2015b, LUZ18.5,
  mirheidari2017toward, mirheidari2018detecting,
  Mirheidari2019dementia, Mirzaei2018, Shinkawa2019, Tanaka2017,
  Weiner2016}, whilst only 27\% have 100 or more participants
\cite{BenAmmar2018, Bertola2014, Clark2016, Fraser2016, Guo2019,
  haider2019assessment, MartinezdeLizarduy2017, orimaye2017predicting,
  PrudHommeaux2011, Prudhommeaux2015, Rochford2012, Troger2018,
  Troger2017, Yu2018}. In fact, 5 report samples with less than 30
participants \cite{EspinozaCuadros2014, Gonzalez-Moreira2015,
  Lopez-de-Ipina2015a, Tanaka2017, Weiner2016}

It is worth noting that those figures represent the dataset size in
full, which is then divided in two, three or four groups, most of the
times unevenly. There are only 6 studies where not only the dataset,
but also each experimental group contains 100 or more
participants/speech samples \cite{BenAmmar2018, Fraser2016, Guo2019,
  haider2019assessment, luz2017longitudinal,
  mirheidari2018detecting}. All of these studies used the \textit{Pitt
  Corpus}.

The \textit{Pitt Corpus} is the largest dataset available. It is used
in full by \citet{BenAmmar2018}, and contains 484 speech samples,
although it is not clear to how many unique participants these samples
belong. With the same dataset, \citet{luz2017longitudinal} reports 398
speech samples, but again, no number of unique participants. However,
another study working with the \textit{Pitt Corpus} does report 473
speech samples from 264 participants \cite{Fraser2016}. It is
important for studies to report numbers of unique participants in
order to allow the reader to assess the risk that the ML models might
actually be simply learning to recognise participants rather than
their underlying cognitive status. This risk can be mitigated, for
example, by ensuring that all samples from each participant are in
either the training set or the test set, but not both.

\subsection*{Data type (table~\ref{tab:data}: Data Details)}
\label{sec:data-type-table}

This column refers to the data used in each reviewed study, indicating
if these data consist of monologues or dialogues, purposefully elicited
narratives or speech obtained through a cognitive test. It also includes
whether data was recorded or recorded and transcribed, and how this
transcription was done (i.e. manual or automatic).

Of the reviewed studies, 82\% used monologue data, and most of them (36) obtained
speech through a picture description task (e.g. \textit{Pitt Corpus}). These
are considered relatively spontaneous speech samples, since participants may describe
the picture in whichever way they want, although the speech content is
always constrained.  Among other monologue studies, eight work with
speech obtained through cognitive tests, frequently verbal fluency
tasks. Only two papers rely on truly spontaneous and natural monologues,
prompted with an open question instead of a picture description
\cite{MartinezdeLizarduy2017, Nasrolahzadeh2018}.

Dialogue data are present less frequently, in 27\% of the studies, and
elicited more heterogeneously. For instance, in structured dialogues
(4 studies), both speakers (i.e. patient and professional) are often
recorded while taking a cognitive test \cite{EspinozaCuadros2014,
  Mirheidari2019dementia, Mirheidari2019computational,
  Tanaka2017}. Semi-structured dialogues (5 studies) are
interview-type conversations where questions are roughly even across
participants. From our point of view, the most desirable data type are
conversational dialogues (5 studies), where interactive speech is
prompted with the least possible constraints \cite{Guinn2014,
  Lopez-de-Ipina2015a, Lopez-de-Ipina2015b, LUZ18.5, Thomas2005}.
A few studies have collected dialogue
data through an intelligent virtual agent (IVA)
\cite{Mirheidari2019dementia, Mirheidari2019computational, Tanaka2017}
showing the potential for data to be collected remotely,
led by an automated computer system.


In terms of data modalities (e.g. audio, text or both), two studies
are the exception where data was directly collected as written text
\cite{fraser2019multilingual, Rentoumi2017}. A few studies (6) work
with audio files and associated ASR transcriptions
\cite{EgasLopez2019,Gosztolya2019, Mirheidari2019computational,
  Sadeghian2017, Toth2018, Troger2017}. Another group of studies (14),
use solely voice recordings \cite{Bertola2014, Chien2018,
  Gonzalez-Moreira2015, Kato2013, Khodabakhsh2015,
  Lopez-de-Ipina2015a, Lopez-de-Ipina2015b, MartinezdeLizarduy2017,
  Meilan2014, Mirzaei2018, Nasrolahzadeh2018, Satt2013, Shinkawa2019,
  Yu2018}. More than half of the studies (55\%) rely, at least
partially, on manually transcribed data. This is positive for data
sharing purposes, since manual transcriptions are usually considered
golden standard quality data. However, methods that rely on
transcribed speech may have limited practical applicability, as they
requires costly and time-consuming, and often (as when ASR is used)
error prone (see section on pre-processing below)
intermediate
steps compared to working directly with the audio recordings.

\subsection*{Other modalities (table~\ref{tab:data}: Data Details)}
\label{sec:other-modal-table}

The most frequently encountered data modality, apart from speech and
language, is structured data related to cognitive examinations, largely dominated
by MMSE and verbal fluency scores. Another modality is video, which is
available in some datasets such as CCC \cite{Guinn2014, LUZ18.5},
AZTITXIKI \cite{Lopez-de-Ipina2015a}, AZTIAHORE
\cite{Lopez-de-Ipina2015b, MartinezdeLizarduy2017}, IVA
\cite{mirheidari2018detecting, Mirheidari2019computational} or the one
in \citet{Tanaka2017}, although it is not included in their
analysis. Other analysed modalities include neuroimaging data, such as
MRI \cite{Clark2016} and fNIRS \cite{Kato2013}, eye-tracking
\cite{Fraser2019, Tanaka2017} or gait information \cite{Shinkawa2019}.

In order to develop successful prediction models for pre-clinical
populations, it is likely that future interactive AI studies will
begin to include demographic information, biomarker data and lifestyle
risk factors \cite{bib:FuenteRitchieLuz2019bmj}.


\subsection*{Data annotation (table~\ref{tab:data}: Data Details)}
\label{sec:data-annot-table}

Group labels and sizes are presented in this section of the Data
Details table, the aim of which is to give information about the
available speech datasets. Accordingly, labels remain as they are
reported in each study, as opposed to the way in which we homogenised
them to describe Comparison Groups in Table~\ref{tab:spicmo}. In other
words, even though the majority of studies annotate their groups as
HC, SCI, MCI and AD, some do not. For example, the HC group is
labelled as CON (control) \cite{Beltrami2016}, NC (normal cognition)
\cite{Bertola2014, Kato2013, Rentoumi2017, Toth2018}, CH (cognitively
healthy) \cite{Chien2018}, and CN (cognitively normal)
\cite{Clark2016}. SCI can also be named SMC \cite{Troger2018}, and
there is a similar but different category (AACD) reported in two other
studies \cite{Weiner2016, Weiner2016a}. MCI and AD are more
homogeneous due to being diagnostic categories that need to meet
certain clinical criteria to be assigned, although some studies do
refer to AD as \textit{dementia} \cite{BenAmmar2018,
  EgasLopez2019}. Another heterogeneous category is CI
(i.e. unspecified cognitive impairment), which is annotated as
\textit{low} or \textit{high} MMSE scores \cite{DArcy2008}, or as
\textit{mild dementia} \cite{Gonzalez-Moreira2015}. \textit{Mild
  dementia} may sound similar to MCI, however the study did not report
diagnostic criteria for MCI to be considered.

This section offers insight into another aspect in which lack of
consensus and uniformity is obvious. Using accurate terminology
(i.e. abiding by diagnosis categories) when referring to each of these
groups could help establish the relevance of this kind of research to 
clinical audiences.

\subsection*{Data balance (table~\ref{tab:data}: Data Details)}

Only 39\% (20) of the reviewed studies present class balance, that is,
the number of participants is evenly distributed across the two, three
or four diagnostic categories \cite{Beltrami2016, BenAmmar2018,
  Chien2018, DosSantos2017, EgasLopez2019, Fraser2019,
  Gonzalez-Moreira2015, Gosztolya2019, Guinn2014,
  haider2019assessment, Kato2013, Khodabakhsh2015,
  MartinezdeLizarduy2017, mirheidari2017toward,
  Mirheidari2019dementia, Mirzaei2018, orimaye2017predicting,
  Rentoumi2017, Roark2011, Tanaka2017}. Among these 20 studies, one
reports only between-class age and gender balance \cite{Tanaka2017},
another one reports class balance, within-class gender balance and
between-class age and gender balance \cite{haider2019assessment}. A
few report balance for all features except for within-class gender
balance, which is not specified \cite{EgasLopez2019, Gosztolya2019,
  Rentoumi2017}. Lastly, there is only one study that, apart from
class balance, also reports gender balance within and between classes,
as well as age and education balance between classes
\cite{fraser2019multilingual}. Surprisingly, nine other studies fail
to report one or more demographic aspects.

Sometimes gender is reported per dataset, but not per class
\cite[e.g.][]{DArcy2008}, and therefore not accounted for in the
analysis, even though is one of the main risk factors
\cite{mielke2014clinical}. Often, \textit{p}-values are appropriately
presented to indicate that demographics are balanced between groups
\cite[e.g.][]{EgasLopez2019}. Unfortunately, almost as often, no
statistical values are reported to argue for balance between groups
\cite[e.g.][]{EspinozaCuadros2014}. There are also cases where where the
text reports demographic balance but neither group distributions nor
statistical tests are presented \cite[e.g.][]{Beltrami2016}. Another
aspect to take into account is the differences between raw and
pre-processed data. For instance, \citet{Lopez-de-Ipina2015a,
  Lopez-de-Ipina2015b} describe a dataset where 20\% of the HC speech
data, but 80\% of the AD speech data, is removed during
pre-processing. Hence, even if these datasets had been balanced before
(they were not) they will definitely not be balanced after
pre-processing has taken place.

It is also worth discussing the reasons behind participant class
imbalance when the same groups are class balanced in terms of
samples. \citet{Fraser2016}, for example, work with a subset of the
\textit{Pitt Corpus} of 97 HC participants and 176 AD participants,
however, the number of samples is 233 and 240, respectively. Similar
patterns apply to other studies where the number of participants and
samples are reported \cite{Guinn2014, Guo2019}. Did HC come for more
visits, or did perhaps AD participants fail to come to later visits or
drop out of the study? These incongruities could be hiding systematic
group biases.

Conclusions drawn from imbalanced data are subject to a greater
probability of bias, especially in small datasets. For example,
certain performance metrics to evaluate classifiers are more robust
(e.g. \textit{F1}) than others (e.g. \textit{acc}) against this
imbalance. Accordingly, in this table, the smaller the dataset, the
more strict we have been when evaluating the balance of its
features. Moving forward, it is desirable that more emphasis is placed
on data balance, not only in terms of group distribution, but also in
terms of those demographic features established risk factors
(i.e. age, gender and years of education).


\subsection*{Data availability (table~\ref{tab:data}: Data Details)}

Strikingly, very few studies make their data available, or even report
on its (un)availability, even when using available data hosted by
a different institution (e.g. studies using the \textit{Pitt
  Corpus}). The majority (77\%, 39 studies) fail to report on data
availability. From the remaining 12 studies, nine use data from DementiaBank
(\textit{Pitt Corpus} or \textit{Mandarin\_Lu}) and do report data
origin and availability. However, only \cite{DosSantos2017,
  orimaye2017predicting} share the exact specification of the subset of \textit{Pitt
  Corpus} used for their analysis, in order for other researchers to
be able to replicate their findings, taking advantage of the
availability of the corpus. The same applies to \citet{LUZ18.5}, who
made available their identifiers for the CCC dataset. One other study,
\citet{Fraser2019}, mentions that data are available upon request
to authors.

\citet{haider2019assessment}, one of the studies working on the
\textit{Pitt Corpus}, has released their subset as part of a challenge
for INTERSPEECH
2020, providing the research community with a dataset matched for age and
gender and with enhanced audio. In such an emerging and heterogeneous field, shared
tasks and data availability are important progression avenues.
\subsection*{Language (table~\ref{tab:data}: Data Details)}

As expected, a number of studies (41\%) were conducted through
English. However, there is a fair amount of papers using data in a
variety of languages, including: Italian \cite{Beltrami2016}, Portuguese
\cite{Bertola2014, DosSantos2017}, Chinese and Taiwanese
\cite{Chien2018}, French \cite{duong2005heterogeneity, Konig2015,
  Mirzaei2018, Troger2017, Troger2018}, Hungarian \cite{EgasLopez2019,
  Gosztolya2019, Toth2018}, Spanish \cite{EspinozaCuadros2014,
  Gonzalez-Moreira2015, Meilan2014}, Swedish \cite{Fraser2019,
  fraser2019multilingual, LundholmFors2018}, Japanese \cite{Kato2013,
  Shinkawa2019, Tanaka2017}, Turkish \cite{Khodabakhsh2015}, Persian
\cite{Nasrolahzadeh2018}, Greek \cite{Rentoumi2017, Satt2013}, German
\cite{Weiner2016a, Weiner2016} or reported as multilingual
\cite{Lopez-de-Ipina2015a, Lopez-de-Ipina2015b,
  MartinezdeLizarduy2017}.

This is essential if screening methodologies for AD are to be
implemented worldwide \cite{bib:DansoTerreraEtAl19apbdar}. The main
caveat, however, is not the number of studies conducted in a
particular language, but the fact that most of the studies conducted
in languages other than English do not report on data availability. As
mentioned, only \citet{DosSantos2017} and \citet{Fraser2019} report their data
being accessible upon request, and \citet{Chien2018} works with data
available from DementiaBank. For speech-based methodology aimed at AD
detection, it would be a helpful practice to make these data
available, so that other groups are able to increase the amount of
research done in any given language.


\subsection*{Pre-processing (table~\ref{tab:methods}: Methodology)}
\label{sec:pre-processing-table}

Pre-processing includes the steps for data preparation prior to data
analysis. It is essential to determine in which shape any given data
is introduced in the analysis pipeline, and therefore, the outcome of
it. However, surprisingly little detail is reported in the
reviewed studies.

Regarding text data, the main pre-processing procedure is
transcription. Transcription may happen manually or through ASR.
The Kaldi speech recognition toolkit 
\cite{madikeri2016implementation}, for instance,  
was used in several recent papers \cite[e.g.][]{EgasLopez2019,
  Mirheidari2019computational}. Where not specified, manual
transcription is assumed. Although many ASR approaches do extract
information on word content \cite[e.g.][]{mirheidari2017toward,
  mirheidari2018detecting, Mirheidari2019dementia, Sadeghian2017,
  Shinkawa2019, Troger2018}, some focus on temporal features, which
are content-independent \cite[e.g.][]{Chien2018, Gosztolya2019}. Some
studies report their transcription unit, that is, word-level
transcription \cite[e.g.][]{Fraser2016}, phone-level transcription
\cite[e.g.][]{Gosztolya2019} or utterance-level transcription
\cite[e.g.][]{Beltrami2016}. Further text pre-processing involves
tokenisation  \cite{Tanaka2017, PrudHommeaux2011, DosSantos2017,
  Chien2018}, lemmatization \cite{fraser2019multilingual} and removal
of stopwords and punctuation \cite{DosSantos2017,
  fraser2019multilingual}. Depending on the research question,
dysfluencies are also removed \cite[e.g.][]{DosSantos2017,
  fraser2019multilingual}, or annotated as relevant for subsequent
analysis \cite[e.g.][]{LundholmFors2018}.

Currently, commercial ASRs are optimised to minimize errors at word
level, and therefore not ideal for generating non-verbal acoustic
features. Besides, it seems that AD patients are more likely to
generate ungrammatical sentences, incorrect inflections and other
subtleties that are not well handled by such ASR systems. In spite of
this, only a few papers, by the same research group, rely on ASR and
report WER (word error rate), DER (diarisation eror rate) or WDER
(word diarisation error rate) \cite{Mirheidari2019dementia,
  mirheidari2018detecting, Mirheidari2019computational}.  It is
becoming increasingly obvious that off-the-shelf ASR tools are not
readily prepared for dementia research, and therefore some reviewed
studies developed their own custom ASR systems \cite{Gosztolya2019,
  Sadeghian2017}.

Regarding acoustic data, pre-processing is rarely reported outside the
audio files being put through an ASR. When reported, it mainly
involves speech-silence segmentation with voice activity deteciton
algorithms (VAD), including segment length and the acoustic criterion
chosen for segmentation thresholds (i.e. intensity)
\cite{haider2019assessment,Kato2013, Konig2015, Lopez-de-Ipina2015b,
  Lopez-de-Ipina2015a, luz2017longitudinal,MartinezdeLizarduy2017,
  Mirzaei2018, Nasrolahzadeh2018, Rochford2012,
  Sadeghian2017,Satt2013, Troger2018, Weiner2016}. It should also
include any audio enhancement procedures, such as volume normalisation
or removal of background noise, only reported in
\citet{haider2019assessment} and \citet{Sadeghian2017}.

We concluded from the reviewed papers that it is not common practice
for authors in this field to give a complete account of the data
pre-processing procedures they followed. As these procedures are
crucial to reliability and replicability of results, we recommend that
further research specify these procedures more thoroughly.

\subsection*{Feature generation (table~\ref{tab:methods}: Methodology)}

Generated speech features are divided into two main groups, text-based
and acoustic features, and follow the taxonomy presented
in Table~\ref{tab:FeatureTaxonomy}. Some studies work with multimodal
feature sets, including images \cite{Tanaka2017} and gait
\cite{Shinkawa2019} measurements.

Text-based features comprise a range of NLP elements, commonly a
subset consisting of lexical and syntactical indices such as
type-token ratio (\textit{TTR}), \textit{idea density} or
\textit{Yngve} and \textit{Frazier} indices. \textit{TTR} is a measure
of lexical complexity, calculated by taking the total number of unique
words, also called lexical items (i.e. types) and dividing by the
total number of words (i.e. tokens) in a given language instance
\cite{biber2007discourse}. \textit{Idea density} is the number of
ideas expressed in a given language instance, with 'ideas' understood
as new information and adequate use of complex propositions. High
early \textit{idea density} seems to be a lower risk predictor for
developing AD later in life, whereas lower idea density appears
associated with brain atrophy \cite{riley2005early}. \textit{Yngve}
\cite{yngve1960model} and \textit{Frazier} \cite{frazier1985} scores
indicate syntactical complexity by calculating the depth of the parse
tree that results from the grammatical analysis of a given language
instance. Both indices have been associated with working memory
\cite{resnik1992left} and showed a declining pattern in the
longitudinal analysis of the written work by Iris Murdoch, a novelist
who was diagnosed with AD \cite{Pakhomov2011}.

In some studies, the research question targets a specific aspect of
language, such as syntactical complexity \cite{LundholmFors2018}, or a
particular way of representing it, such as speech graph attributes
\cite{Bertola2014}. \citet{Fraser2016} present a more comprehensive
feature set, including some acoustic features. Similar to
\citet{Fraser2016}, although less comprehensive, a few other studies
combine text-based and acoustic features \cite{Beltrami2016,
  Gonzalez-Moreira2015, Guo2019, mirheidari2017toward,
  Mirheidari2019dementia, Roark2011, Sadeghian2017, Shinkawa2019,
  Tanaka2017, Troger2018}. However, most published research is
specific to one type of data or another.

The most commonly studied acoustic features are prosodic temporal
features, which are almost invariably reported, followed by ASR-related
features, specifically pause patterns. There is also focus on
spectral features (features of the frequency domain
representation of the speech signal obtained through application of
the Fourier transform), which include \textit{MFCCs}
\cite{EgasLopez2019}. The most comprehensive studies include spectral,
ASR-related, prosodic temporal, voice quality features \cite{Guo2019,
  Khodabakhsh2015, Lopez-de-Ipina2015a, MartinezdeLizarduy2017,
  Meilan2014, Mirheidari2019dementia, Mirzaei2018}, as well as
features derived from the Higuchi Fractal Dimension
\cite{Lopez-de-Ipina2015b} or from higher order spectral analysis
\cite{Nasrolahzadeh2018}. It is worth noting here that
\citet{Tanaka2017} extract $F_0$'s coefficient of variation per
utterance. The decision to not extract $F_0$'s mean and SD was due to
their association with individual differences and sex. Similarly,
\citet{Gonzalez-Moreira2015} report $F_0$ and functionals in
semitones, because research argues that using semitones to express
$F_0$ reduces gender differences \cite{t1981differential}, which is
corroborated by the choice of semitones in the standardised eGeMAPS
\cite{haider2019assessment}.

Studies using spoken dialogue recordings extract turn-taking patterns,
vocalisation instances and speech rate \cite{LUZ18.5,
  Tanaka2017}. Those focusing on transcribed dialogues also extract
turn-taking patterns, as well as dysfluencies
\cite{mirheidari2017toward, Mirheidari2019dementia,
  Mirheidari2019computational}. \citet{Guinn2014} work with
longitudinal dialogue data but do not extract specific dialogue or
longitudinal features.

With regards to feature selection, 30\% of the studies do not report
feature selection procedures. Amongst those that do, the majority
(another 30\%) report using a filter approach based on a statistical
index of feature differences between classes, such as
\textit{p}-values, Cohen's \textit{d}, \textit{AUC} or
\textit{Pearson's} correlation. Others rely on wrapper methods
\cite{Mirzaei2018}, RFE \cite{mirheidari2017toward,
  Mirheidari2019dementia}, filter methods based on information gain
\cite{BenAmmar2018, Nasrolahzadeh2018}, PCA \cite{Kato2013}, best
first greedy algorithm \cite{Sadeghian2017}, and cross-validation,
seeking through the iterations for which feature type contributes more
to the classification model \cite{Lopez-de-Ipina2015b}.

Despite certain similarities and a few features being common to most
acoustic works (i.e. prosodic temporal), there is striking
heterogeneity among studies. Since they usually obtain features using
\textit{ad hoc} procedures, these studies are seldom comparable,
making it difficult to ascertain the state-of-the-art in terms of
performance, as pointed out before, and assess further research
avenues. However, this state of affairs may be starting to change as
the field matures. \citet{haider2019assessment}, for instance, chose
to employ standardised feature sets (i.e. ComPare, eGeMAPS, emobase)
obtained through formalised procedures \cite{eyben2010opensmile} which
are extensively documented and can be easily replicated. Furthermore,
one of these feature sets, eGeMAPS, was developed specifically to
target affective speech and underlying physiological
processes. Utilising theoretically informed, standardised feature sets
increases the reliability of a study, since the same features have
been previously applied (and can continue to be applied) to other
engineering tasks, always extracted in the exact same way. Likewise,
we argue that creating and utilising standardised feature sets will
improve this field by allowing cross-study comparisons. Additionally,
we recommend that the approach to feature generation should be more consistently
reported to enhance study replicability and generalisability.


\subsection*{ML task/method (table~\ref{tab:methods}: Methodology)}
\label{sec:ml-taskmethod-table}

Most reviewed papers employ supervised learning, except for
a study that uses cluster analysis to investigate distinctive discourse
patterns amongst participants \cite{duong2005heterogeneity}.

As regards choice of ML methods, very few papers report the use of
artificial neural networks \cite{Beltrami2016, BenAmmar2018},
recurrent neural networks \cite{Chien2018}, multi-layer perceptron
\cite{Clark2016, Lopez-de-Ipina2015a, Lopez-de-Ipina2015b,
  mirheidari2017toward, Sadeghian2017} or convolutional neural
networks \cite{MartinezdeLizarduy2017, mirheidari2018detecting}. This
is probably due to the fact that most datasets are relatively small,
and these methods require large amounts of data.  Rather, most studies
use several conventional ML classifiers, most commonly SVM, NB, RF and
\textit{k}-NN and then compare their performance. Although these
comparisons must be assessed cautiously, a clear pattern seems to
emerge with SVM consistently outperforming other classifiers.

Cognitive scores, particularly MMSE, are available with many datasets,
including the most commonly used, \textit{Pitt Corpus}. However, these
scores mostly remain unused except for diagnostic group assignments,
or more rarely, as baseline performance \cite{Prudhommeaux2015,
  Sadeghian2017, Shinkawa2019}, in studies that conclude that MMSE is
not more informative than speech based features. All supervised
learning approaches work towards classification and no regression over
cognitive scores is attempted. We regard this as a gap that could be
explored in future research.

It is worth noting, however, that some attempts at prediction of MMSE
score have been presented in workshops and computer science
conferences that are not indexed in the larger biobliography
databases. These approaches achieved some degree of success. 
\citet{linz2017predicting}, for instance, trained a regression model that
used the SVF to predict MMSE scores and obtained a mean absolute error
of 2.2. A few other works used the \textit{Pitt Corpus} for similar
purposes, such as \citet{al2017detecting}, who extracted 811 acoustic
features to build a regression model able to predict MMSE scores with
an average mean absolute error of 3.1; or \citet{pou2018learning}, who
used a multiview embedding to capture different levels of cognitive
impairment and achieved a mean absolute error of 3.42 in the
regression task. Another publication with the \textit{Pitt Corpus} is
authored by \citet{yancheva2015using}, who extracted a more
comprehensive feature set, including lexicosyntactic, acoustic, and
semantic measures, and used them to predict MMSE scores. They trained
a dynamic Bayes network that modeled the longitudinal progression
observed on these features and MMSE over time, reporting a mean
absolute error of 3.83. This is, actually, one of the very few works
attempting a progression analysis over longitudinal data.


\subsection*{Evaluation techniques (table~\ref{tab:methods}: Methodology)}
\label{sec:eval-techn-table}

A substantial proportion of studies (43\%) do not present a baseline
against which study results can be compared. Amongst the remaining
papers, a few set specific results from a comparable work as their
baseline \cite{luz2017longitudinal, Nasrolahzadeh2018} or from their
own previous work \cite{orimaye2017predicting}. Others calculate their
baseline by training a classifier with all the generated features,
that is, before attempting to reduce the feature set with either
selection or extraction methods \cite{BenAmmar2018,
  EspinozaCuadros2014, Toth2018}, with cognitive scores only
\cite{Fraser2019, Prudhommeaux2015, Sadeghian2017, Shinkawa2019} or by
training a classifier with demographic scores only
\cite{Gosztolya2019}. Some baseline classifiers are also trained with
a set of speech-based features that excludes the feature targeted by
the research question. Some examples are studies investigating the
potential of topic model features \cite{fraser2019multilingual},
emotional features \cite{Lopez-de-Ipina2015a}, fractal dimension
features \cite{Lopez-de-Ipina2015b}, higher order spectral features
\cite{Nasrolahzadeh2018} or feature extracted automatically, as
opposed to manually
\cite{mirheidari2018detecting,PrudHommeaux2011,Troger2018}. Some
studies choose random guess or naive estimations (ZeroR)
\cite{haider2019assessment,LUZ18.5,Rentoumi2017, Thomas2005,
  Weiner2016a} as their baseline performance.

While several performance metrics are often reported,
\textit{accuracy} is the most common one. While it seems
straightforward to understand a classifier's performance by knowing
its \textit{accuracy}, it is not always appropriately informed. Since
\textit{accuracy} is not robust against dataset imbalances, it is only
appropriate when diagnostic groups are balanced, such as when reported
in \citet{Khodabakhsh2015, Roark2011}. This is especially problematic
for works on imbalanced datasets where accuracy is the only metric
reported \cite{DArcy2008, EspinozaCuadros2014, Fraser2016, Guo2019,
  MartinezdeLizarduy2017, Meilan2014, Mirheidari2019computational,
  Sadeghian2017, Thomas2005, Troger2017}. Clinically relevant metrics
such as \textit{AUC} and \textit{EER} \cite[e.g.][]{Konig2015,
  Satt2013}, which summarise the rates of false alarms and false
negatives, are reported in less than half of the reviewed studies.

Cross-validation (CV) is probably the most established practice for
classifier evaluation. It is reported in all papers but five, of which
two are not very recent \cite{DArcy2008, Thomas2005}, another two do
not report CV but report using a hold-out set \cite{Beltrami2016,
  Chien2018}, and only one reports using neither CV nor a hold-out set
procedure \cite{BenAmmar2018}. There is a fair amount of variation
within the CV procedures reported, since datasets are limited and
heterogeneous. For example, leave-one-out CV is used in one third of
the reviewed papers, as an attempt to mitigate the potential bias
caused by using a small dataset. Several other studies choose
leave-pair-out CV instead \cite{Fraser2019, orimaye2017predicting,
  PrudHommeaux2011, Prudhommeaux2015, Roark2011, Yu2018}, since it
produces unbiased estimates for \textit{AUC} and also reduces
potential size bias.  There is also another research group who attempted to
reduce the effects of their imbalance dataset by using stratified CV
\cite{Weiner2016, Weiner2016a}. Lastly, no studies report hold-out set procedures, except for the two
mentioned above, with training/test sets divided at 80/20\% and
85/15\%, respectively, and another study where the partition percentages
are not detailed \cite{Yu2018}.

There is a potential reporting problem in that many studies
do not clearly indicate whether their models' hyper-parameters were
optimized on the test set within or outside each fold of the
CV. However, CV is generally considered the best method of evaluation
when working with small datasets, where held-out set procedures would
be even less reliable, since they would involve testing the system on
only a few samples. CV is therefore an appropriate choice for the
articles reviewed. The lack of systematic model validation on entirely
separate datasets, and the poor practice of using accuracy as
the single metric in imbalanced datasets, could compromise the
generalisability of results in this field. While it is worth
noting that the former issue is due to data scarcity, and therefore
more difficult to address, a more appropriate selection of performance
metrics could be implemented straight away to enhance the robustness
of current findings.

\subsection*{Results overview (table~\ref{tab:methods}: Methodology)}
\label{sec:results-overv-table}

Performance varies depending on the metric chosen, the type of data
and the classification algorithm used. Hence, it is very difficult to
summarise these results. The evaluated classifiers range between 50\%
or even lower in some cases, up to over 90\%
accuracy. However, as we have pointed out, performance figures must be
interpreted with caution due to the potential biases introduced by
dataset size, dataset imbalances and non standardised \textit{ad hoc}
feature generation. . Since these biases cannot be fully accounted for
and models are hardly comparable to one another, we do not think it is
meaningful to further highlight the best performing models. Such
comparisons will become more meaningful when all conditions for
evaluation can be aligned, such as in the ADReSS challenge
\cite{luz2020ADReSS}, which provides a benchmark dataset (balanced and
enhanced) and commits to a reliable study comparison.

Further research on the methodology and how different algorithms
behave with certain types of data will shed light on why some
classifiers perform even worse than random while others are close to
perfect. This could simply be because the high performing algorithms
were coincidentally tested on 'easy' data (e.g. better quality,
simpler structures, very clear diagnoses), but the problem could also
classifier specific and therefore differences would be associated with
the choice of algorithm.  Understanding this would influence the
future viability of this sort of technology.




\subsection*{Research implications (table~\ref{tab:clinical}: Clinical applicability)}
\label{sec:rese-impl-table}

This section reviews the papers in terms of novelty, replicability and
generalisability, three aspects key to future research.

As regards \textbf{novelty}, the newest aspect of each research paper
is succinctly presented in the tables. This is often conveyed by the title of an
article, although caution must be exercised with regards to how this information is
presented. For example, \citeauthor{Troger2018}'s title (2018) reads
``Telephone-based Dementia Screening I: Automated Semantic Verbal
Fluency Assessment'', but only when you read the full text does it
become clear that such telephone screening has been simulated.

There is often novelty in pilot studies, especially those presenting
preliminary results for a new project, hence involving brand new data
\cite{Beltrami2016, Lopez-de-Ipina2015b} or tests for a newly
developed system or device \cite{MartinezdeLizarduy2017}. Outside of
those, assessing novelty in a systematic review over a 20-year span
can be complicated --- what was novel 10 years ago might
not be novel today. For example, 3-way classification entailed novelty
in \citeAD{Bertola2014}, as well as 4-way classification did in
\citeAD{Thomas2005} with text data and little success, and later in
\citeAD{Nasrolahzadeh2018} with acoustic data and an improved
performance. Given its low frequency and its naturalness, we have
chosen to present the use of dialogue data \cite{Guinn2014, LUZ18.5,
  Khodabakhsh2015, mirheidari2018detecting, Tanaka2017, Weiner2016} as
a novelty relevant for future research. Other examples of novelty
consist of automated neuropsychological scoring, either by automating
traditional scoring \cite{Bertola2014} or by generating a new type of
score \cite{Clark2016, Prudhommeaux2015}.

Methodological novelty is also present. Even though most studies apply
standard machine learning classifiers to distinguish between
experimental groups, two approaches do stand out:
\citeauthor{duong2005heterogeneity}'s (2005) unique use of cluster
analysis (a form of unsupervised learning) with some success, and the
use of ensemble \cite{Clark2016, DosSantos2017} and cascaded
\cite{Fraser2019} classifiers, with much better results. Some studies
present relevant novelty for pre-processing, generating their own
custom ASR systems \cite{Gosztolya2019, Sadeghian2017, Satt2013,
  Toth2018}, which offers relevant insight about off-the-shelf
ASR. While this is based on word accuracy, some of the customized ASR
systems are phone-based \cite{Gosztolya2019, Toth2018} and seem to
work better with speech generated by participants with AD. Another
pre-processing novelty is the use of dynamic threshold for pause
behaviour \cite{Rochford2012}, which could be essential for
personalised screening. With regards to feature generation, ``active
data representation'' is a novel method utilized in conjunction with
standardised feature sets by \citet{haider2019assessment}, who
confirmed the feasibility of a useful tool that is open software and
readily available (i.e. ComParE, emobase and eGeMAPS). A particularity
of certain papers is their focus on emotional response, analysed from
the speech signal \cite{Lopez-de-Ipina2015a,
  Lopez-de-Ipina2015b}. This could be an avenue for future research,
since there are other works presenting interesting findings on
emotional prosody and AD
\cite{haiderfuente2020affective,horley2010emotional}. Last, but not
least, despite the mentioned importance of early detection, most
papers do not target early diagnosis, or do it in conjunction with
severe AD only (i.e. if the dataset contains participants at different
stages). Consequently, \citeAD{LundholmFors2018} introduced a crucial
novelty by not only assessing, but actively recruiting and focusing on
participants at the pre-clinical stage of the disease (SCI).


Another essential novelty is related to longitudinal aspects of data
\cite{Troger2017, Weiner2016, Yu2018}. The vast majority of studies
work on monologue cross-sectional data, although some datasets do include
longitudinal information (i.e. each participant has produced several
speech samples). This is sometimes discarded, either by treating each
sample as a different participant, which generates subject dependence
across samples \cite{Weiner2016a}; or by cross-observation averaging,
which misses longitudinal information but does not generate this
dependence \cite{orimaye2017predicting,Yu2018}. Other studies
successfully used this information to predict change of cognitive
status within-subject \cite{Pakhomov2014, Weiner2016}.
\citet{Guinn2014} work with longitudinal dialogue data that becomes
cross-sectional after pre-processing (i.e. they conglomerate samples by the same
participant) and they do not extract specific dialogue features.

The novelty with most clinical potential is, in our view, the
inclusion of different types of data, since something as complex as AD
is likely to require a comprehensive model for successful
screening. However, only a few studies combine different sources of
data, such as MRI data \cite{Clark2016}, eye-tracking
\cite{Fraser2019}, and gait \cite{Shinkawa2019}. Similarly, papers where
human-robot interaction \cite{mirheidari2018detecting,
  Mirheidari2019dementia, Mirheidari2019computational, Tanaka2017} or
telephone-based systems \cite{Troger2018, Yu2018} are implemented also
offer novel insight and avenues for future research. These approaches
offer a picture of what automatic, cost-effective screening could look
like in a perhaps not so distant future.

On a different front, \textbf{replicability} is assessed based on
whether the authors report complete and accurate procedures of their
research. Replicability has research implications because, before
translating any method into clinical practice, its performance needs
to be confirmed by other researchers being able to reproduce similar
results. In this review, replicabilility is labelled as \textit{low}, \textit{partial} and \textit{full}. 
When we labelled an article as \textit{full} with regards to
replicability, we meant that their methods section was considered to
be thorough enough to be reproduced by an independent researcher, from
the specification of participants demographics and group size to the
description of pre-processing, feature generation, classification and
evaluation procedures. Only three articles were labeled as \textit{low}
replicability \cite{mirheidari2018detecting, Thomas2005, Weiner2016a},
as they lacked detail in at least two of those sections (frequently
data information and feature generation procedures). Twenty-two and
twenty-five studies were labelled as \textit{partial} and \textit{full},
respectively. The elements most commonly missing in the \textit{partial}
papers are pre-processing \cite[e.g.][]{EspinozaCuadros2014} and
feature generation procedures \cite[e.g.][]{Roark2011}, which are
essential steps in shaping the input to the machine learning
classifiers. It must be highlighted that all \textit{low} replicability papers
are conference proceedings, where text space is particularly
restricted. Hence, it does not stand out as one of the key problems of
the field, even though it is clear that the description of
pre-processing and feature generation must be improved.

The last research implication is \textbf{generalisability},
which is the degree to which a research approach may be attempted with
different data, different settings or real practice. Since
generalisability is essentially about how translatable research is,
most aspects in this last table are actually related to it:
\begin{itemize}
\item Whether \textit{external validation} has been attempted is
  directly linked to generalisability;
\item \textit{feature balance:} results obtained in imbalanced
  datasets are less reliable and therefore less generalisable to other
  datasets;
\item \textit{contextualization of results:} for something to be
  generalisable is essential to know where it comes from and how does it compare to similar research;
\item \textit{spontaneous speech}: speech spontaneity is one aspect of
  naturalness, and the more natural the speech data, the more
  representative of "real" speech and the more generalisable;
\item \textit{conversational speech}: we propose that conversational
  speech is more representative of "real" speech;
\item \textit{content-independence:} if the
  classifier input includes features that are tied up with task
  content (e.g. lexical, syntactic, semantic, pragmatics), some degree
  of generalisability is lost;
\item \textit{Transcription-free:} a model that needs no transcription
  is free from ASR or manual transcription constraints, relying only
  on acoustic features. We suggest this to increase generalisability,
  for example, by being language-independent, therefore facilitating
  method usability with non-English language for which corpus training
  is less feasible due to even more severe data
  scarcity. Transcription-free methods also facilitate the protection
  of users' privacy, as they do not focus on speech content, which
  could encourage ethics committees to reduce restrictions on data
  collection, thereby addressing data scarcity.
\end{itemize}

Just as replicability, it is labelled as \textit{low},
\textit{moderate} and \textit{high}, depending of how many of the
aforementioned criteria each study meets. Different to what we
described with regards to replicability, the majority of studies (20)
are labelled with \textit{low} generalisability, 17 as
\textit{moderate} and 14 as \textit{high}. The most common reasons for
decreased generalisability are dependence on content, followed by
dependence on ASR or other transcription methods, although the two are
related. Content-dependence makes it difficult to apply to other tasks
or data \cite[e.g.][]{Beltrami2016,Mirheidari2019computational}. This
is even more pronounced in those studies where the approach heavily
relies on word content, such as \textit{n}-grams
\cite[e.g.][]{orimaye2017predicting}. Linguistic models that target
only one linguistic aspect are also \textit{low} generalisability,
particularly if this aspect is language-dependent
\cite[e.g. syntaxis][]{LundholmFors2018}.  Examples of \textit{high}
generalisability include models relying solely on acoustic features,
therefore free of content and transcription constrains
\cite[e.g.][]{Gonzalez-Moreira2015}, and especially if a standardised
available feature set is used \cite{haider2019assessment}. Other
generalisable studies present more than one dataset
\cite[e.g.][]{EgasLopez2019}, different languages in the same study
\cite[e.g.][]{fraser2019multilingual}, conversational data
\cite[e.g.][]{LUZ18.5}, a system designed for direct real application
\cite[e.g.][]{Tanaka2017} and/or data from real scenarios
\cite{Mirheidari2019dementia}.

\subsection*{Clinical potential (table~\ref{tab:clinical}: Clinical
  applicability)}
\label{sec:clin-potent-table}

The clinical applicability table aims to directly assess whether
reviewed research could be translatable into clinical
practice. Generalisability (discussed above) is essential for this
purpose, but it will not be included here to avoid redundancy.  We
also note that clinical applicability of a diagnostic test is a
somewhat vague construct in that one might need applicability in a
clinical population {\em or} applicability for a clinician to
understand its use.  From a clinician's perspective, the translational
steps from research on speech and language biomarkers to clinical use
are not unlike those of any other diagnostic tool.  This highlights,
as we point out in the conclusion,
that this
translational development pathway would benefit from joint development
between clinicians, speech and language experts, and AI experts
working in concert. The other systematic aspects chosen to evaluate
clinical potential are:

\begin{itemize}
\item \textbf{External validation:} in the majority of studies,
  data are collected detached from clinical practice and later analysed
  for result reporting.  The majority of papers (84\%) present neither external
  validation procedures nor a system design that involves them. Only
  four studies, all of them by the same group
  \cite{mirheidari2017avatar, mirheidari2018detecting,
    Mirheidari2019dementia, Mirheidari2019computational}, collect
  their data in a real life setting (doctor-patient
  consultations). Another four studies take into account feasibility
  for clinical screening within their system design, for example,
  collecting data directly with a computerised decision support system
  \cite{MartinezdeLizarduy2017}, through human-robot interaction
  \cite{Tanaka2017}, a computer-supported screening tool
  \cite{Troger2017}, or simulating telephone-based data
  \cite{Troger2018}.
    
\item \textbf{Potential application:} 78\% of the reviewed papers
  present a method that could be applied as a diagnosis support system
  for MCI \cite[e.g.][]{fraser2019multilingual} or AD
  \cite[e.g.][]{Fraser2016}. The remaining studies work on disease
  progression by including SCI participants \cite{LundholmFors2018},
  predicting within-subject change \cite{Weiner2016} or discriminating
  between HC, MCI and AD stages \cite[e.g.][]{Konig2015}.
    
\item \textbf{Global Health:} although this could include a broad
  range of aspects, for the purpose of this review we have chosen to
  mention the language of the study and the processing unit of
  choice. This is because most research is done in
  English (41\%), and work published in other languages helps towards
  methods being more universally applicable. Also, because 
  smaller the processing units (i.e. phoneme vs. word),
  tend to be more generalisable across languages. The most common
  processing unit is the sentence (63\%), followed by conversations
  (16\%), words (8\%), syllables (4\%) and phonemes (4\%).
    
\item \textbf{Remote application:} for such a prevalent disease,
  remote screening could significantly reduce the load on health
  systems. The majority of the studies, 67\%, do not mention the
  possibility of their method being used remotely or having being
  designed for remote use, and only 25\% suggest this as a possibility when
  motivating their project or discussing the results. Only four studies
  (2\%), actually bring this into practice by experimenting with
  multi-modal human-robot interaction \cite{Tanaka2017},
  infrastructure-free \cite{Troger2017} or telephone-based
  \cite{Troger2018, Yu2018} approaches.
\end{itemize}

A further aspect, not explicitly included on this table is {\em model
  interpretability}. While the accepted opinion is that the
clinicians' ability to be able to interpret an AI model is essential
for the adoption of AI technologies in medicine, the issue is still
the subject of lively debate, with influential machine learning
researchers like Geoff Hinton arguing that ``clinicians and regulators
should not insist on explainability''
\cite{bib:WangKaushalKhullar19shcd}.  In terms of biomarkers of
disease, very few if any clinicians understand the fine detail of an
MRI report; it is the results presented to them that clinicians
contextualise rather than the statistical or AI journey these results
have been on to be presented to them. It could be argued that the case
of speech and language biomarkers is no different. Of the papers
reviewed here, only 4 mention interpretability or model interpretation
explicitly \cite{Fraser2019,Fraser2016,Bertola2014,Yu2018}. However,
inherently interpretable models are used in a number of studies. Such
interpretable methods were indicated in the above section on AI methods
and
include: linear regression, logistic regression, generalised linear
and additive models, decision trees, decision rules, RuleFit, naive
Bayes and K-Nearest neighbors \cite{bib:Molnar19inml}, and in some
cases linear discriminant analysis. As shown on
Table~\ref{tab:methods}, 57\% of the studies reviewed included at
least one of these types of models in their evaluation, even though
most such inclusions were made for comparison purposes.

As regards the selected criteria, the result tables highlight that
research undertaken using non-English speech data almost invariably
includes acoustic features, either as part of a larger feature set,
such as \citet{Beltrami2016} in Italian; or exclusively relying on
acoustic features, such as \citet{Nasrolahzadeh2018} in Persian,
\citet{Weiner2016} in German, \citet{EspinozaCuadros2014,
  Gonzalez-Moreira2015, Lopez-de-Ipina2015b, Meilan2014} in Spanish
and \cite{Kato2013} in Japanese.  Apart from English, only Portuguese
and Chinese have been researched exclusively with text features
\cite{Bertola2014, DosSantos2017, Chien2018}.

Some of the field's needs clearly arise here. Firstly, there is a need
for actual attempts to use these models in real clinical practice. For
twenty years, conclusions and future directions of these research
papers have suggested this, but very few published studies do bring it
into a realisation. Secondly, there is a need for enhanced focus on
disease progression and risk prediction. Most studies mention the need
for AD to be diagnosed earlier than it is now, and yet not many do
actually work in that direction. Thirdly, further investment on
research performed on languages other than English is needed, and
increased focus on smaller language units, which are more
generalisable to other languages or other samples of the same
language. Alternatively, we suggest that a shift towards acoustic
features only would potentially address these difficulties. Finally,
one of the most obvious advantages of using Artificial Intelligence
for cognitive assessment is the possibility of using less
infrastructure and less personnel. In order for this to become a
reality, the remote applicability of these methods requires more
extensive research.


\subsection*{Risk of bias (table~\ref{tab:clinical}: Clinical applicability)}
\label{sec:risk-bias-table}

This column highlights sources of potentially systematic errors or
other circumstances that may introduce bias in the inferred
conclusions. These can be summarised as follows:

\begin{itemize}
\item \textbf{Feature balance:} class, age, gender and education
  balances are essential for experimental results to be unbiased. Only
  13 studies (25\%) are balanced for these main features, and another
  five are balanced in terms of class but not in terms of other
  features. In the studies that seek to address class imbalance in
  their datasets, the main strategies used are subsampling
  \cite{MartinezdeLizarduy2017,haider2019assessment}, use of
  statistical methods such as stratified CV \cite{Weiner2016a},
  and careful 
  choice of evaluation methods including use of the UAR metric
  \cite{Gosztolya2019} and
  ROC curve analysis \cite{Yu2018}. 
\item \textbf{Suitable metrics:} equally important for bias prevention
  is choosing the right performance metrics to evaluate machine
  learning classifiers. For example, with a class-imbalanced dataset,
  accuracy is not a robust metric and should therefore not be used, or
  at least, complemented with other measures. However, 18
  studies (35\%) working with an imbalanced dataset report accuracy
  only.
\item \textbf{Contextualized results:} referring mainly to whether the
  reported research is directly and quantitatively compared to related
  works, or, ideally, whether a baseline against which results can be
  compared is provided. Only 61\% of the studies reviewed provide such
  context.
\item \textbf{Overfitting:} studies would apply both CV and
  held-out sets to ensure their models do not over-fit. CV should be
  applied when tuning ML hyper-parameters when training the model, and
  the held-out set should be used to test the model on strictly unseen
  data. The majority of the studies do report CV (78\%), but even more
  studies (90\%) do not report hold-out set. Hence, there is a high
  risk that the reviewed models are, to some degree, overfitted to the
  data they have been trained with. Ideally, models should also be
  validated on entirely separate datasets. Only one of the studies
  reviewed carries out this kind of validation, although their method aims to use speech alignment in order to automatically score a cognitive task, instead of investigating the potential for dementia prediction of the linguistic or speech features themselves \cite{Prudhommeaux2015}.
\item \textbf{Sample size:} labelled as up to 50 participants
  ($ds\leq50$), up to 100 participants ($ds\leq100$) or over 100
  participants ($ds>100$). The results show that 13 studies were
  carried on smaller datasets (i.e. $ds\leq50$), 24 studies carried on
  medium-sized datasets (i.e. $ds\leq100$) and 14 studies carried on
  modestly larger datasets (i.e. $ds>100$). However, seven of the
  studies carried on a medium-sized dataset and one study carried on a
  larger dataset attempted 3-way or even 4-way
  classification. Therefore, the group sizes of these studies are
  further reduced by the fact that the original dataset size needs to
  be divided into three or four groups, instead of the two groups used
  for binary classification.
\end{itemize}

We decided to use these numerical labels to classify the datasets,
instead of assigning categories such as small or large, because even
the largest dataset of the reviewed studies is relatively small when
put into a machine learning context. All in all, there is a clear need
for larger available datasets that are also balanced in terms of class
and main risk factors. On larger datasets, it should be more
straightforward to increase methodological rigour (e.g. by using CV,
hold-out sets) and to seek for more active and systematic ways to
prevent overfitting.


\subsection*{Strengths/Limitations (table~\ref{tab:clinical}: Clinical
  applicability)}
\label{sec:strengthsl-table-ref}

In our view, a few desirable qualities should be present in AI
research for AD, in order for it to be finally translatable into
clinical practice. These are:

\begin{itemize}
\item \textbf{Spontaneous speech:} we consider spontaneous speech data
  to be more representative of real life spoken language. Although
  speech data obtained through non-spontaneous, constrained cognitive
  tasks present methodological advantages, we argue that spontaneous
  speech is desirable for cognitive monitoring due to its ubiquity,
  naturalness and relative ease of collection. Under this criterion,
  we seek not only to explore the advantages of using speech for
  cognitive screening, but also the suitability for continuous and
  longitudinal collection. 65\% of the papers meet this criterion with
  this by using open question data (e.g. free episodic recalls,
  discourses prompted by a picture, conversational dialogues). The
  remaining papers rely on contrained data, obtained for example by
  recording the words produced in a fluency test.
\item \textbf{Conversational speech:} similarly, we deem
  conversational speech to be more representative of real life spoken
  language than monologue speech. Here again we find a trade-off
  between naturalness and standardisation. While monologueas are
  easier to handle (by requiring fewer preprocessing steps) and may
  avoid potential confunding factors present in dialogues
  (e.g. relationships between speakers, conversational style, cultural
  norms surrounding doctor-patient conversations), some methods may
  take advantage of these very factors for cognitive screening as they
  enrich the cognitive mechanisms involved in the interaction
  \citep{Pickering2004}. Of the reviewed papers, only 18\% report the
  use of dialogue (i.e. structured, semi-structured or
  conversational).
\item \textbf{Automation:} most of the reviewed papers claim some
  degree of automation in their procedure, but looking closely, only
  37\% describe a fully (or nearly fully) automatic method, from
  transcription to classification. Another 37\% describe a partially
  automatic procedure, frequently automating feature generation and/or
  classification steps, but with a manual transcription and/or manual
  feature set reduction. The rest describe methods that require manual
  interference at almost every stage, and were therefore deemed to not
  be automatic.
\item \textbf{Content-independence:} this is desirable in order for
  successful methods to be equally successful when speech is elicited
  in different ways (i.e. with different tasks, which imply different
  content). 55\% of the papers report procedures that do rely on
  content-related characteristics of speech, such as word content. The
  rest either rely solely on acoustic features or phoneme based
  transcribed features, unrelated to word content.
\item \textbf{Transcription-free:} as mentioned above, ASR methods are
  an automatic alternative to manual transcription, but they are not
  free of constrains. Therefore, we consider transcription free
  approaches to offer a more relevant contribution to the clinical
  application of AI for AD detection. Under this criterion, 35\% of
  the reviewed papers use a transcription-free approach, whereas the
  rest rely on either manual or ASR transcriptions.
\end{itemize}

Only two studies meet all five criteria with a "yes"
\cite{Khodabakhsh2015, LUZ18.5}.  In our view, the field needs to
further explore the use spontaneous speech (ideally conversational),
and indeed we have observed renewed interest in its use during the
time span of this review, as AI becomes increasingly involved, as
shown in Figure~\ref{fig:records}. Automation also needs to be pursued
by trying to bridge the gaps where automation becomes challenging,
namely, during transcription, as well as during feature generation and
feature set reduction (i.e. feature selection and feature
extraction). Seeking automation entails a complex trade-off, since
there is clearly valuable information about a person's cognitive
status reflected in the content of what they say, as well as how they
say it and how they choose to structure it. In addition, not all
linguistic features are content-dependent and metrics such as word
frequency, vocabulary richness, repetitiveness and syntactic
complexity are not linked to semantic content or meaning. However,
processing language to obtain these metrics makes automation and
generalisation more difficult, specially as regards non-English
data. While content-related information can offer insights into the
nature of the disease and its development, reviewing the potential for
AI systems in terms of practical usefulness in clinical settings for
cognitive health monitoring requires considerations of
content-independent and transcription-free approaches due to their
ease of implementation, successful performance and more
straightforward generalisability.

\subsection*{Overall Conclusions}
\label{sec:conclusions}

We have conducted the first systematic review on the potential
application of interactive artificial intelligence methods to AD detection and
progression monitoring using natural language processing and speech technology to
extract ``digital biomarkers'' for ML modelling. 

Given the somewhat surprising quantity and variety of studies we
found, it seems reasonable to conclude that this is a very promising
field, with potential to gradually introduce changes into clinical
practice. Almost all studies report relatively high performance,
despite the difficulties inherent to the type of data used and the
heterogeneity of the methods. When compared to neuropsychological
assessment methods, speech and language technology were found to be at least
equally discriminative between different groups. It is worth noting
that the most commonly used neuropsychological test, MMSE, has been
criticised \cite{carnero2014should} due to its inherent biases and
lack of sensitivity to subtle symptoms. In this context,
interactive AI could offer the same or better performance as a
screening method, with the additional advantages of being implemented
automatically and, possibly, remotely. Notwithstanding, while most of
the papers hereby reviewed highlight the potential of AI and ML
methods, no actual translation into clinical practice has been
achieved. One might speculate that this slow uptake, despite nearly 20
years of research in this field, is due to difficulties in attaining
meaningful interdisciplinary cooperation among between AI/ML research
experts and clinicians. We expect that the growing interest in and
indeed adoption of AI/ML methods in medicine will provide the stimulus
needed for effective translation to take place in this field as it has
in others.  
Despite an unexpectedly high number of records found eligible to
review (51), the field remains highly heterogeneous with respect to
the available data and methodology. It is difficult to compare results
on an equal footing when their conclusions are drawn from monologue,
dialogue, spontaneous and non-spontaneous speech data. Similarly,
different choices of processing units (varying from phoneme and
syllable to a word, sentence or a longer instance) pose additional
comparability challenges. Furthermore, while machine learning
methodology is somewhat standardised through a wealth of open-source
tools, the feature generation and feature set reduction procedures are
not. Feature generation varies greatly, with the same feature falling
into slightly different categories depending on the
study. Consequently, abiding by a standard taxonomy like the one
proposed by \citet{voleti2019review}, which we adapted in Table
\ref{tab:FeatureTaxonomy}, becomes essential in order to make
cross-study comparisons. Surprisingly, many studies do not report on
their approach to feature set reduction, or do it very vaguely, giving less than
enough detail for replication. To our knowledge, only one study
\cite{haider2019assessment} relies on standardised feature sets
available to the research community, while all other articles extract
and calculate speech and language indices in an \textit{ad hoc},
non-consensual way.

Furthermore, although cross-validation is implemented in most
publications as an evaluation technique, many studies proceed with
feature set reduction outside a cross-validation setting. That is,
both training and testing data are used to find the relevant features
that will serve as input to the classifier input. Additionally,
although it is standard practice to tune machine learning models using
a preferred performance metric (i.e. \textit{acc, EER, AUC, F1}), we
must recognise the potential effect this might have on the reliability and
generalisability of such models. If CV is done correctly (i.e. not optimizing hyper-parameter tuning within the test set of each fold), the models created in any given fold of the CV procedure are tested on unseen data, although many studies do not provide this information. Barely any of the reviewed studies
reported a hold-out set procedures or experiments on an entirely separate dataset, which would be the ideal scenario for robust model validation. 

One of the reasons behind this lack of rigour is the size and variable
quality of the datasets, which prevents adequate subsets to be
generated while the size and integrity of the experimental groups is
maintained. Consequently, we are confident that establishing certain
standards on data and methodology will also increase the strictness of
study evaluation. With regards to data type and availability, firstly,
we should mention that data collection in this field is particularly
difficult due to ethic constraints, due to the personally-identifying
nature of speech data. Secondly, a benchmark dataset is essential to
set the long overdue baselines of the field. Such baselines should not
only refer to detection performance for SCI, MCI and AD classes, but
also to regression models able to predict cognitive scores, which is
repeatedly proposed but hardly ever done, and prediction of
progression and risk. Thirdly, we note that conversational dialogue
(i.e. natural dialogue) is an under-explored resource in this
field. As noted before, although monologue data presents
methodological advantages, dialogue data has the potential to offer
richer results precisely due to factors that under certain
methodological frameworks might be dismissed as confounds. That is, an AI
system trained to evaluate speakers interaction, cultural norms and
conversational styles has potential to be more versatile 
in monitoring cognitive health for different people, in different
settings and at different times of the day. Furthermore, dialogue data
could be easier and more natural to collect in real life (i.e. we
spend part of our day interacting with somebody else), as well as more
representative of a broader range psycholinguistic aspects such as
alignment and entrainment at different linguistic levels
\cite{Pickering2004}, which might be relevant to AD detection.

With regards to methodology, we recommend a wider use of standardised
feature sets, such as eGeMAPS \cite{eyben2016geneva}, purposefully
developed to detect physiological changes in voice
production. Needless to say, other feature sets should also be built
and tested systematically, for the field to move toward finding a
golden standard. Further benefits of a consensual set of features
entail the possibility of tracking those features ``back to the
brain'', in order to find their neural substrate and hence
contributing to knowledge of the neuropathology of AD.

In terms of aims and objectives, research suggests that embedded
devices installed in the home to monitor patient health and safety may
delay institutionalisation \cite{fredericks2018cal}, and therefore
more emphasis should be placed on the feasibility of remote
evaluations. To this end, we propose that future research focuses on
natural conversations, which are straightforward to collect passively,
continuously and longitudinally in order to monitor cognitive health
through an AI system. We also argue that focusing on cohorts already
diagnosed with AD is no longer a relevant task for AI. As noted
earlier, the majority of studies reviewed in this paper focus on
diagnosis. We argue that emphasis should shift towards earlier stages
of the disease, when pre-clinical unobserved changes start. Future
research should therefore attempt to include healthy populations at
higher risk of developing AD in larger scale longitudinal studies, as
well as compare those populations to lower risk populations.  There is
good potential for interactive AI technology to contribute at those
stages, given its increasingly ubiquitous presence in our lives,
through wearable devices, smartphones, ``smart homes'', and ``smart
cities''.

In addition, novel AI/ML digital biomarkers
\cite{bib:CoravosKhozinMandl19d} could be used in combination with
established biomarkers to target populations at risk of later dementia
onset, as has already been proposed
\cite{bib:FuenteRitchieLuz2019bmj}. It needs to be emphasised that
recorded data are considered personal data (i.e. with potential to
identify a subject), with the ethical and regulatory hurdles this
entails as regards data collection and analysis. We suggest that the
field would benefit from revised ethics agreements to facilitate
speech data collection, as well as from data sharing across
institutions until datasets reach sufficient size to support complex
machine learning structures and results are robust enough to encourage
clinical applications. Increased collaboration between clinicians and
AI experts should favour these developments.

\section*{Acknowledgements}
We thank Marshall Dozier, the Academic Support Librarian for her help with the search queries and the PROSPERO protocol. This research was supported by the Medical Research Council (MRC), grant number MR/N013166/1.

\section{Conflict of Interest/Disclosure Statement}
The authors have no conflict of interest to report. 



\bibliographystyle{ios-abbrvnat}           
\bibliography{main}
%

\clearpage
\section*{Supplementary material}

\subsubsection*{Keys to table interpretability}
The following conventions and acronyms were adopted when reviewing the
research articles.

When a conference paper was later extended and published in a
different venue, only the latter publication was included. Percentages
were rounded to the nearest decimal place.  Education is expressed in
average years, same as age, unless otherwise specified. For the
purposes of this review we do not distinguish between acoustic and
paralinguistic features, and the tables we will always designate these
features as ``acoustic''. Where the type of classifier or model used
is not specified on the table, this means it was not reported in the
paper reviewed.
Since ``data balance'' can be understood in, at least, three different
ways, we created the following acronyms to standardise and simplify the
description of this characteristic on the tables:
\begin{itemize}
\item \textbf{dataset Feature Balance (FB):} where the "F" is replaced
  for the initial of each particular feature. This refers to whether a
  certain feature, e.g. gender, is evenly distributed in the dataset
  as a whole. Consider, for instance, a dataset with 100 participants,
  60 healthy controls (HC), 30 of whom are female, and 40 AD
  participants, 30 of whom are female. As regards gender, this is
  indicated in the table as ``No-GB'' (no gender balance) because the
  ratio of female to male participants in this dataset is
  60:40. Regarding class, this dataset would also be "No-CB" (no class
  balance), because the ratio of HC to AD is also 60:40A. Age and
  education are reported as a between class features only.
    
\item \textbf{Within-Classes Feature Balance (WCFB):} This indicates
  whether a certain feature, such as gender, is evenly distributed
  {\em within}
  each class. In the above described sample dataset, this would be
  indicate as ``HC:
  WCGB, AD: no-WCGB'', because the gender ratio is 30:30 in HC, but 30:10
  in AD. This type of balance makes sense for gender, since it is a
  category within the classes. However, age and education are
  generally reported as group averages, and hence it is not possible to
  report their balance within class.
    
\item \textbf{Between-Classes Feature Balance (BCB: F):} This denotes
  whether a feature is evenly distributed {\em across} experimental
  classes. Our example dataset would be described as ``BCB: no-G'',
  because the number of female is indeed balanced (30 in both groups),
  but the number of males is not (30 in HC vs 10 in AD). Again, his
  type of balance makes sense for gender, since it is a category
  within the classes. However, age and education are generally
  reported as group averages, and hence balance between classes is
  equivalent to their balance across the dataset. Since they are
  reported to control for their potential confounding effect between
  groups, we will report their between-class balance.
    
\end{itemize}

Gender balance is reported as GB, WCGB, BCB: G, or,
alternatively, no-GB, no-WCGB, BCB: no-G. Class balance is
reported as CB or no-CB. Age and education, with respect to group
average are reported within BCB: A, E or BCB: no-A, no-E. No-BCB
indicates that none of the features are balanced between classes,
whereas BCB (without further specification) indicates that all
three are.  Lastly, some studies report the number of speech or text
samples as well as the number of participants per group and then take
only one of those figures for analysis. In these cases class balance
will be indicated followed by an \textit{m} or an \textit{n} depending
on whether the comparison groups are balanced in terms of samples or
participants, respectively. For example, CB\textit{m} would indicate
class balance based on number of samples per class, whereas
no-CB\textit{n} would indicate class imbalance in terms of number
participants per class, both of which could coexist in the same
study. A similar logic applies to other types of balance
(e.g. no-WCGB\textit{m}, BCB\textit{m}: no-A, G, no-E). When it is
unspecified whether the analyses and reported results are based on
number of participants or number of samples, this will be indicated as
"unclear".

\subsection*{Abbreviations and Acronyms}\label{subsec:abbr}

\begin{table}[htb]
\caption{Diagnostic abbreviations}
  \label{tab:dx+terms}
\footnotesize
\centering
\begin{tabular}{ ll  } 
\hline
\multicolumn{2}{c}{\textbf{Diagnostic Groups}} \\
 \hline
AD  & Alzheimer's Disease or Dementia \\ 
CI  & Cognitive Impairment (unspecified)\\ 
HC  & Healthy Controls \\ 
MCI & Mild Cognitive Impairment \\ 
SCI & Subjective Cognitive Impairment \\ \hline
\end{tabular}
\begin{tablenotes}{\footnotesize
\item Note: For the purpose of this review, labels such as ``normal
  elderly'' or ``cognitively normal'' are noted as HC; dementia groups as
  AD. For the pre-clinical stage, Subjective Memory Loss (SML) is
  equated to SCI. CI refers to the symptomatic group where no official
  diagnosis term is stated.}
\end{tablenotes}
\end{table}

\begin{table}
  \caption{General textual abbreviations}
  \label{tab:textualabbrev}
  \footnotesize
  \begin{tabular}{ ll}
    \hline
\multicolumn{2}{c}{\textbf{General terms and noun phrases}} \\
    \hline
ast             & assessment                                              \\ 
avail           & available                                               \\ 
B/L             & baseline                                                \\ 
corr            & correlation                                             \\ 
demogr          & demographics                                            \\ 
ds              & dataset                                                 \\ 
ft              & features                                                \\ 
ibid            & see previous footnote on same dataset/paper/topic       \\ 
incl            & included                                                \\ 
info            & information                                             \\ 
lex             & lexical                                                 \\ 
m               & number of samples                                       \\ 
meas            & measurement (s)                                         \\ 
n               & number of participants                                  \\ 
NI              & Neuroimaging                                            \\ 
pp              & participants                                            \\ 
rec             & recording                                               \\ 
repr            & representation                                          \\ 
S/N             & average sentences per narrative or sample               \\ 
seg             & segmentation                                            \\  
syl             & syllable                                                \\ 
tr              & transcript                                              \\ 
utt             & utterance                                               \\ 
w/              & with                                                    \\ 
w/o             & without                                                 \\ 
W/S             & average words per sentence or sample                    \\ \hline
\end{tabular}
\end{table}
%
\begin{table}[h!tb]
\caption{Methods and metrics}
  \label{tab:abbrMethods+Metrics}
\footnotesize
\begin{tabular}{ ll  } 
 \hline
 \multicolumn{2}{c}{\textbf{Methods}} \\
 \hline
ADR             & Active Data Representation                              \\ 
ASR             & Automatic Speech Recognition                            \\ 
CNN             & Convolutional Neural Network                            \\ 
CV              & Cross-validation                                        \\ 
DR              & Data representation                                     \\ 
DT              & Decision Trees                                          \\ 
GC              & Gaussian Classifier                                     \\ 
GNB             & Gaussian Naive Bayes                                    \\  
HOS             & Higher Order Spectral (analysis)                        \\                                  
IG              & Information Gain                                        \\ 
\textit{k}-NN   & \textit{k}-Nearest Neighbour \\
LDA             & Linear Discriminant Analysis \\  
LASSO (LR)      & Least Absolute Shrinkage and Selection Operator         \\ 
LM              & Language Model                                          \\ 
LR              & Logistic Regression                                     \\ 
LOO (CV)        & Leave-one-out                                           \\ 
LPO (CV)        & Leave-pair-out                                          \\ 
LSA             & Latent Semantic Analysis                                \\ 
LSTM (RNN)      & Long-short-term-memory                                  \\ 
MLP             & Multi-layer Perceptron                                  \\ 
NN              & Neural Network                                          \\ 
PCA             & Principal Components Analysis                           \\ 
RBF (SVM)       & Radial Basis Function (kernel)                          \\ 
RF              & Random Forest                                           \\ 
RFE             & Recursive Feature Elimination                           \\ 
RNN             & Recurrent Neural Network                                \\ 
SGD             & Stochastic Gradient Descent                             \\ 
SVM             & Support Vector Machine                                  \\ \hline
 \multicolumn{2}{c}{\textbf{Metrics}} \\
 \hline
ACC             & Accuracy                                                \\ 
AUC             & Area Under the Curve                                    \\ 
CER             & Classification Error Rate                               \\ 
EER             & Equal Error Rate                                        \\ 
FA              & False Alarms                                            \\ 
MLU             & mean length of utterance                                \\ 
N               & noun frequency                                          \\  
pc              & Precision                                               \\ 
rc              & Recall                                                  \\ 
sp              & Specificity                                             \\ 
ss              & Sensitivity                                             \\ 
ROC             & Receiving operating characteristic (curve)              \\ 
TP              & True Positives                                          \\ 
UAR             & Unweighted Average Recall                               \\ 
V               & verb frequency                                          \\ \hline
\end{tabular}
\end{table}


\clearpage
\onecolumn
   \section{SPICMO (PICOS) table}
The first table is based on the PICOS design, widely used in the
clinical field. Its columns are:
\begin{itemize}
\item \textbf{\underline{P}opulation}: total number of participants
  followed by number of participants per group (always starting with
  the less impaired group). Average demographic figures (i.e. age,
  education, MMSE) follow the same order.
\item \textbf{\underline{I}nterventions}: assessments that
  participants underwent as part of the study. These are usually
  either cognitive or full clinical assessments, recorded speech tasks
  and written tasks.
\item \textbf{\underline{C}omparison groups}: different stages of
  cognitive impairment of the study participants conform the groups to
  be compared. These are Healthy Controls (HC), Subjective Cognitive
  Impairment (SCI), Mild Cognitive Impairment (MCI), Alzheimer's
  Disease (AD) and Cognitive Impairment (CI), when unspecified. This
  terminology is not standardised across publications, but we have
  standardised it for the purpose of this review. Hence, for instance,
  normal controls (NC), healthy elderly (HE), Subjective Memory
  Complaints (SMC) or Dementia (e.g. Alzheimer's Type Dementia) are
  hereby equated to HC, SCI and AD, respectively.
\item \textbf{\underline{O}utcomes of interest}: detection, prediction
  or discrimination performance of the method used in an article. This
  mostly includes classification metrics, such as overall accuracy,
  sensitivity and specificity.
\item \textbf{\underline{S}tudy aim/design}: most frequently,
  automatic detection of a target group when compared to a healthy
  one, or automatic discrimination between different stages of target
  groups. It also includes the main design, i.e. text vs. speech,
  narrative vs. monologue.
\end{itemize}

We have extended this by adding a column on methods, an
essential part of this review:

\begin{itemize}
\item \textbf{\underline{M}ethodology}: brief overview of the approach
  for feature generation (i.e. acoustic analysis or natural
  language processing), as well as the approach for feature set reduction (when reported). These are feature selection (i.e. filtering or wrapping) and feature extraction (i.e. combination or transformation of original features, e.g. PCA, LSA, ADR). This section also mentions the machine learning task used in the paper (i.e. machine learning task).
\end{itemize}

Lastly, we considered it to be more intuitive for this review to have
information about \textbf{S}tudy aim at the beginning, and therefore,
the conventional order of the columns has been shifted, yielding
SPICMO (study aim, population, intervention, comparisons, methodology
and outcomes) as a result.

\onecolumn
\begin{landscape}
\setlength{\footskip}{3cm}
\scriptsize
\newcolumntype{R}[1]{>{\raggedright\arraybackslash}p{#1}}
\setlength\LTleft{-2.2cm}
\begin{longtable}{R{1.5cm}R{2.8cm}R{3.3cm}R{2.9cm}R{2.8cm}R{3.7cm}R{3.0cm} }
  \caption{\label{tab:spicmo} SPICMO (PICOS) table} \\\toprule
  \textbf{Study} & \textbf{Study aim/design} & \textbf{Population} & \textbf{Interventions} & \textbf{Comparison groups} & \textbf{Methodology} & \textbf{Outcomes of interest}  \\\midrule
  \endfirsthead	
  \caption{SPICMO (PICOS) table (ctd.)} \\\toprule
  \textbf{Study} & \textbf{Study aim/design} & \textbf{Population} & \textbf{Interventions} & \textbf{Comparison groups} & \textbf{Methodology} & \textbf{Outcomes of interest}  \\\midrule
  \endhead	

\citet{Beltrami2016}
&
Automatic detection of MCI based on acoustic and linguistic fts from narrative speech data. 
&
39 pps: 20 HC, 19 MCI. Aged range: 50-75 years old; educ: high school/university. Italy.
&
Cognitive ast.\break
\textbf{Speech task:} narratives picture description, working day, last dream.
&
HC and MCI. Based on cognitive ast (MMSE, MoCA, GPCog, CDT, VF, CANTAB-PAL)
&
Acoustic and linguistic analysis of speech for ft extraction, statistics for ft selection, ML for group classification.
&
Detection performance: 76.9\% accuracy (picture task) ($F_1=78.1\%$).
\\

\citet{BenAmmar2018}
&
Automatic detection of AD based on linguistic fts from narrative speech data.
&
484 samples: 242 HC, AD; unreported pps from \underline{Pitt}\footnote{This paper reports the number of speech samples they used, but not to how many pps they belonged to.}. Age $>$ 44; educ $>$ 7; MMSE $>$ 10. USA.
&
Clinical ast. \break
\textbf{Speech task:} narrative picture description (Cookie Theft). 
&
HC and MCI. Based on cognitive ast (i.e. MMSE).
&
Audio enhancement, linguistic analysis for ft extraction, ML for ft selection and group classification.
&
Detection performance: 79\% accuracy.
\\

\citet{Bertola2014}
&
Discrimination between HC, MCI and AD using graph analysis on word sequences obtained from SVF data.
&
100 pps: 25 HC, aMCI, a+mdMCI, AD. Median age: 76, 76 (MCI), 79; educ: 4, 4, 4; MMSE: 27, 25, 20. Brazil.
&
Clinical ast (i.e. medical, cognitive).\break
\textbf{Speech task:} recorded SVF answers (animals).
&
HC, amnesic single/multiple domain (a/+mdMCI), and AD. Based on cognitive ast (Katz, Lawton, MMSE).
&
Graph analysis for ft extraction from word sequences, statistics for ft selection (incl SVF scores), ML for group classification.
&
Discrimination performance: $AUC=0.68$ HC-MCI, $AUC=0.73$ MCI-AD, $AUC=0.88$ HC-AD (graph attributes only).\\

\citet{Chien2018}
&
Automatic detection of AD based on acoustic fts from narrative speech and SVF data.
&
60 pps: 30 HC,  30 AD from Mandarin\_Lu\footnote{\underline{Mandarin\_Lu} corpus is hosted within DementiaBank}. 150 speech samples, demographics unreported. China, Taiwan.
&
\textbf{Speech task:} recorded SVF answers (fruits, locations) and narrative picture description.
&
HCand AD. Unreported criteria.
&
Manual acoustic analysis for ft extraction, ML for group classification.
&
Detection performance: $AUC=0.95$.
\\

\citet{Clark2016}
&
Automatic prediction of MCI conversion to AD from automatic SVF scores combined with neuroimaging data. 
&
107 pps: 83 MCI-non (46F), 24 MCI-con (15F). Avg age: 68.7, 73.8; educ: 16, 16; MMSE: 27.9, 25.1. USA.
&
Cognitive ast, brain MRI.\break
\textbf{Speech task:} recorded SVF (vegetables, animals) and OVF (letters F, A, S).
&
MCI-non and MCI-con (upon conversion to AD). Based on Petersen criteria (incl. MMSE, CDR).
&
Automatic scoring of verbal fluency answers (electronically transcribed) and neuroimaging scores for ft extraction, ML for group classification.
&
Conversion prediction performance (at 4-year follow-up): $AUC=0.872$ with automatic scores.
\\

\citet{DArcy2008}
  &
Detection of probable CI based on manually and automatically extracted temporal speech fts.
  &
87 pps: 50 HC 37 probable CI. Age range: 62-92, 62\%F, MMSE over 24 (excl severe CI). Ireland.
  &
Cognitive ast .\break
\textbf{Speech task}: narrative picture descriptions,  Scribe\footnote{5 pictures from an in-house task (Picture Taboo). Scribe consists of sentences designed to cover English language phones.}, SVF (animals), word list, Heidi passage.
  &
HC (MMSE $>27$) and probable CI (MMSE $\leq27$).
  &
Manual and automatic temporal analysis of speech for ft extraction (syntactic, acoustic), ML for group classification.
 &
Detection performance: 76.74\% accuracy (manual approach far superior than automatic).Vowel 17\% longer in probable CI.\\

\citet{DosSantos2017}
&
Automatic detection of MCI based on speech fts from transcribed narratives in English and Portuguese.
&
86 \underline{Pitt}\footnote{86 pps: 43 HC (20F), MCI (16F). Avg age: 64.1, 69.3; educ over 7, MMSE over 10.} pps. USA.\break
40 \underline{Cinderella}\footnote{40pps: 20 HC (16F), MCI (14F). Avg age: 74.8, 73.3; educ: 11.4, 10.8.} pps. USA.\break
43 \underline{ABCD}\footnote{43 pps: 20 HC, 23 MCI. Avg age: 61, 72; educ: 16, 13.3.} pps. Brazil. 
&
Cognitive ast.\break
\textbf{Speech task:} picture description (\underline{Pitt}), story retelling (Cinderella), recall (ABCD).
&
HC and MCI. Based on cognitive ast (\underline{Pitt}), Petersen criteria (\underline{Cinderella}), clinical diagnosis (\underline{ABCD}).
&
Topological network and linguistic analysis for ft extraction, ML for group classification.
&
Detection performance: 65\% \underline{Pitt} accuracy, 65\% \underline{Cinderella}, 75\% \underline{ABCD}.
\\

\citet{duong2005heterogeneity}
  &
Description of discourse patterns and heterogeneity in AD based on transcribed narrative speech.
  &
  99 pps: 53 HC (40F), 46 AD (39F). Avg age: 73.8, 74.3; educ: 10.2, 8.3. Canada.
  &
  Cognitive ast. \break
\textbf{Speech task:} P1 and P7 narrative picture descriptions from PENO\footnote{PENO is a cognitive battery in French \cite{joanette1995evaluation}. "Bank robbery" and "Car accident" are language subtests from this battery.}.
  &
 HC (NE: normal elderly) and AD. Based on NINCDS-ADRDA criteria \cite{mckhann1984clinical}.
  &
Discourse analysis of the transcribed narratives, cluster analysis to group pps with similar discourse patterns.
  &
 Clusters inconclusive for prototypical AD discourse (heterogeneity). 4 different discourse patterns for P1 and 5 for P7.\\

\citet{EgasLopez2019}
&
Discrimination between HC, MCI and AD using the i-vector approach on acoustic fts from narrative speech.
&
75 pps: 25 HC, MCI, AD. Avg age: 73.96, 72.4, 70.72; educ: 10.76, 10.84, 12.08; MMSE: 29.24, 27.16, 23.92.
&
Cognitive ast.\break
\textbf{Speech task:} narrative of previous day, immediate and delayed recall of 2 short films.
&
HC, MCI and AD. Based on cognitive ast (i.e. MMSE, CDT, ADAS-Cog)
&
Acoustic analysis of speech recordings ft extraction, i-vector approach for dimensionality reduction, ML for group classification. 
&
Discrimination performance: 56\% accuracy and $F_1=78.4\%$, all tasks.\break 
$F_1=79.2\%$ immediate recall only.
\\

\citet{EspinozaCuadros2014}
&
Automatic detection of MCI based on speech fts from interviews and narratives.
&
19 pps: 11 HC (6F), 8 MCI (2F). Avg age: 78.9, 80.3; educ: 8, 5
&
\textbf{Speech task:} structured interview recorded from MEC\footnote{Mini-Examen Cognoscitivo (MEC) is the Spanish adaptation of the MMSE\cite{pena2012evaluacion}.} and a reading short passage.\footnote{In particular, a Spanish version of "The Grandfather Passage"\citet{darley1975motor}.}.
&
HC and MCI.\break
Unreported criteria.
&
Acoustic analysis of speech recordings for ft extraction, statistics for ft selection, ML for group classification.
&
Detection performance: 78.9\% accuracy with seven prosodic features from the passage reading task.
\\

\citet{Fraser2019}
&
Automatic detection of MCI based on multi-\ modal fts from language tasks (audio, text, eye-tracking).
&
55 pps: 29 HC (21F), 26 MCI ( 14F). Avg age: 67.8, 70.6; educ: 13.3, 14.3; MMSE: 29.6, 28.2.
&
Cognitive ast. \break
\textbf{Speech task:} picture description (Cookie Theft), read short text\footnote{Short texts obtained from the International Reading Speed Texts (IReST).} aloud and in silence. 
&
HC and MCI. Based on Petersen criteria.\break
Gothemburg MCI Study.
&
Multi-modal approach for ft extraction, cascaded ML for group classification (ft, mode, task and session).
&
Detection performance: 83\% accuracy (AUC$=0.88$), with multi-modal fts. \break
84\% accuracy (AUC$=0.90$) incl cognitive scores.
\\

\citet{fraser2019multilingual}
&
Automatic detection of MCI based on topic modelling and information content in Swedish and English.
&
In-domain\footnote{Data including MCI pps: 67 Gothenburg. 36 HC (23F), 31 MCI (16F). Avg age: 67.9, 70.1; educ: 13.1, 14.1; MMSE: 29.6, 28.2.}: 67 pps, \underline{Gothenburg} ds, Sweden.\break
Out-of-domain\footnote{Data not including MCI pps: 96 Karolinska (52F) and 78 \underline{Pitt} (48): all HC. Avg age: 57.2, 63.9; educ: 13, 13.9; MMSE: N/A, 29.1.}: 96 pps, \underline{Karolinska} ds Sweden; 78 pps, \underline{Pitt}, USA.
&
Cognitive ast.\break
\textbf{Speech task:} picture description (Cookie Theft) spoken or written (Karolinska ds).
&
HC and MCI. Based on cognitive ast and clinical diagnosis.
&
Linguistic analysis for ft extraction, multilingual topic models for dimensionality reduction and ft seletion, ML for group classification. 
&
Detection performance: \break
63\% accuracy in English; \break
72\% accuracy in Swedish.\break
Based on information content.
\\

\citet{Fraser2016}
  &
Automatic detection of AD based on linguistic (mostly) and acoustic fts from narrative speech.
  &
473 samples\footnote{\underline{Pitt}: repeated samples from 264 participants (97 HC, 167 AD).}: 233 HC (151F), 240 AD (158F). Avg age: 65.2, 71.8; educ: 14.1, 12.5, MMSE: 29.1, 18.5.
  &
Clinical ast (i.e. medical, cognitive). \break
\textbf{Speech task:} narrative picture description (Cookie Theft). 
  &
HC and AD. Based on cognitive ast (i.e. MMSE).
  &
Acoustic and linguistic analysis of speech for ft extraction, factor analysis for dimensionality reduction, ML for group classification.
  &
  Detection performance: 81.92\% accuracy.
  Factors identified (4): semantic, acoustic, syntactic and information.\\

\citet{Gonzalez-Moreira2015}
&
Automatic detection of mild dementia based on acoustic fts from narrative speech in Spanish.
&
20 pps: 10 HC (1F), 10 CI\footnote{This category is named MD (mild dementia) in study.} (4F). Avg age 78.9, 80.3; educ 7.8, 4.
&
Cognitive ast.\break
\textbf{Speech task:} read short text aloud ("The Grandfather Passage".
&
HC and CI. Based on cognitive ast (i.e. MEC scores).
&
Acoustic analysis for ft extraction of recorded speech, ML for group classification.
&
Detection performance: 85\% accuracy based on four prosodic fts (articulation rate, mean syllables duration, F0 sd and mean).
\\

\citet{Gosztolya2019}
&
Automatic detection of AD and MCI, as well as discrimination between HC, MCI and AD based on acoustic (ASR) and linguistic speech fts.
&
75 pps: 25 HC, MCI, AD. Avg age: 70.72, 72.4, 73.96; educ: 12.08, 10.84, 10.7; MMSE: 29.24, 27.16, 23.92.\break
225 recordings.
&
Cognitive ast.\break
\textbf{Speech task:} narrative of previous day, immediate and delayed recall of 2 short films.
&
HC, MCI and AD groups. Based on cognitive ast (i.e. MMSE, CDT, ADAS-Cog).
&
Acoustic (ASR) and linguistic analysis for ft extraction, ML for group classification.
&
Detection performance: 86\% HC-AD, 80\% HC-MCI, 81.3\% HC-CI accuracy.\break
Discrimination performance: 66.7\% (morphologic and acoustic). 
\\

\citet{Guinn2014}
&
Automatic detection of AD based on linguistic fts from dialogue transcripts.
&
56 pps from \underline{CCC}\footnote{Carolina Conversations Collection (CCC) \cite{pope2011finding} is conversational corpus consisting of conversations about health and healthcare gathered longitudinally with people with different chronic conditions, including AD.}: 28 nonAD, 28 AD. Multiple transcripts per pp: 204 nonAD, 77 AD. 
&
\textbf{Speech task:} conversational interview about pp's chronic condition and their experience in healthcare.
&
HC (nonAD: patients with chronic conditions unrelated to AD) and AD. Based on clinical diagnosis.
&
Linguistic analysis for ft extraction of dialogue transcripts (syntax, semantics, pragmatics), ML for gorup classification.
&
Detection performance: $Prec_{AD}=80.8\%$, $recall_{AD}=0.75\%$; \break
$Prec_{nonAD}=79.3\%$, $recall_{nonAD}=82.1\%$.
\\

\citet{Guo2019}
&
Automatic detection of AD based on linguistic fts from narrative speech.
&
268 pps\footnote{subset from \underline{Pitt} with multiple samples per participant. This study used 498 samples: 242 HC, 256 AD.}: 99 HC (58F), 169 AD (114F). Avg age: 61.3, 71; educ: 13.3, 11.8; MMSE: 27.9, 18.7. 
&
Clinical ast (i.e. medical, cognitive). \break
\textbf{Speech task:} narrative picture description (Cookie Theft). 
&
HC and AD. Based on cognitive ast (i.e. MMSE).
&
Linguistic analysis for ft extraction (phonetics, semantics, syntax, pragmatics) including perplexity, ML for group classification. 
&
Detection performance: 85.4\% accuracy including perplexity fts derived from language models.
\\

\citet{haider2019assessment}
&
Automatic detection of AD based on standardised acoustic ft sets extracted from narrative speech.
&
164 pps (\underline{Pitt}): 82 HC (46F), AD (46F). Aged 50-80, mostly 65-75 (BWGB); educ over 7 years: MMSE over 10. 
&
Clinical ast (i.e. medical, cognitive). \break
\textbf{Speech task:} narrative picture description (Cookie Theft). 
&
HCand AD. Based on cognitive ast (i.e. MMSE).
&
Acoustic analysis for ft extraction from the recorded narratives, ADR\footnote{Active Data Representation: novel method presented in this paper.} for ft selection and representation, ML for group classification.
&
Detection performance: 71.34\% accuracy with one ft set; 78-80\% accuaracy with "hard fusion" of all ft sets.
\\

\citet{Kato2013}
&
Discrimination between HC, MCI and AD with a two-phase system based on speech fts cerebral blood flow.
&
48 pps: 20 HC (13F), 19 MCI (13F), 9 AD (4F). Age range: 64-92 years old. Other demographics unreported.
&
Cognitive ast.\break
\textbf{Speech task:} topics hometown and childhood, HDS-R, memory tasks.\break
Simultaneous fNIRS\footnote{Functional near-infrared spectroscopy: measures cortical brain activity by monitoring changes of oxy/deoxygenated hemoglobin concentration.}.
&
HC, MCI and AD. Based on cognitive ast. (i.e. CDR = 0, 0.5 or 1).
&
Multivariate statistics to generate cognitive rating (SPCIR) based on prosody, ML for group classification in two phases: fNIRS and prosody.
&
Discrimination performance: 85.4\% overall accuracy. \break
32\% MCI participants misclassified into HC group.\\

\citet{Khodabakhsh2015}
&
Automatic detection of AD based on acoustic fts from conversational speech. 
&
54 pps: 27 HC (15F), 27 AD (10F). Age range: 60-80 years old. Other demographics unreported.
&
\textbf{Speech task:} pps were asked casual questions to elicit 10 min of spontaneous conversation.
&
HC and AD. Unreported criteria.
&
Voice activity detection (VAD) and acoustic analysis for ft extraction from recordings, ML for group classification.
&
Detection performance: 
79.2\% with best pair of fts (log of voicing ratio + avg absolute delta pitch.
\\

\citet{Konig2015}
&
Discrimination between HC, MCI and AD based on automatically extracted speech fts across different tasks.
&
\underline{Dem@care}: 64 pps.\break 15 HC (9F), 23 HC (12F), 26 AD (13F). Avg age 72,73,80; educ:\footnote{Mode, that is, most frequent educational category.}: uni, col, hs; MMSE: 29,26,19.
&
Cognitive ast.\break
\textbf{Speech task:} countdown, picture description, repetition, SVF (animals).
&
HC, MCI and AD. Based on subjective memory complain (HC), Petersen criteria (MCI), NINCDS-ADRDA (AD)
&
Acoustic analysis for ft extraction from speech recordings, statistics for ft selection, ML for group classification.
&
Detection performance:
$EER_{HC-MCI}=21\%$,
$EER_{HC-AD}=13\%$,
$EER_{MCI-AD}=20\%$.\break
Equal ss-sp: 79\%, 87\%, 80\%
\\

\citet{Lopez-de-Ipina2015b}
&
Discrimination between HC and stages of AD based on acoustic fts, incl. emotional response (pilot study).
&
10 pps (\underline{AZTITXIKI}\footnote{Subset of \underline{AZTIAHORE}, which is, in turn, a subset of \underline{AZTIAHO}.}):\break 
5 HC/AD (2F).\break
AD: 1ES, 2SS, 2AS. Age label: middle (HC), elderly (HC and AD).
&
\textbf{Speech task:} telling pleasant stories, recounting pleasant feelings, conversational interaction.
&
HCand ES, IS, AS, which stand for early, intermediate and advanced AD. \break
Unreported criteria.
&
Acoustic analysis and emotional response analysis (ERA) for ft extraction, ML for group classification.
&
Discrimination performance:
93.79\% accuracy with speech and emotional fts. Unclear whether this result is on 4 groups or 2.
\\

\citet{Lopez-de-Ipina2015a}
&
Discrimination between HC and stages of AD based on acoustic fts, incl. emotional response.
&
40 pps (\underline{AZTIAHORE}\footnote{which is, in turn, a subset of AZTIAHO}. 20 HC (10F), 20 AD (12F). AD: 4ES, 10SS, 6AS. Age range: 20-98 HC, 68-98 AD. Others unreported.
&
\textbf{Speech task:} telling pleasant stories, recounting pleasant feelings, conversational interaction.
&
HC\footnote{Group annotated as CR (control group) in the paper, equated to HC for the purpose of this review.} and ES, IS, AS, which stand for early, intermediate and advanced AD. \break
Unreported criteria.
&
Acoustic analysis and fractal dimension for ft extraction, incl. emotional response, ML for group classification.
&
Discrimination performance: accuracy reported per class, avg. 96.89\%. Overall performance unclear.
\\

\citet{LundholmFors2018}
&
Discrimination between HC, SCI and MCI based on syntactic fts extracted from narrative speech.
&
\underline{Gothemburg}\footnote{Gothenburgh MCI data set \citet{wallin2016gothenburg}}: 90pps. \break36 HC (23F), 23 SCI (14F), 31 MCI (16F). Avg age: 67.9, 66.3, 70.1; educ: 13.2, 16.1, 14.1; MMSE: 29.6, 29.5, 28.2.
&
Clinical ast (i.e. medical, cognitive). \break
\textbf{Speech task:} narrative picture description (Cookie Theft). 
&
HC, SCI and MCI. Based on clinical diagnosis.
&
Linguistic analysis for syntactic ft extraction, statistical analysis for feature selection, and ML for group classification.
&
Discrmination performance (binary detection):\break
HC-MCI: $F_1 = 0.68$,
HC-SCI: $F_1 = 0.54$,
SCI-MCI: $F_1 = 0.66$\\

\citet{luz2017longitudinal}
&
Automatic detection of AD based on acoustic fts extracted directly from voice recordings of narrative speech.
&
398 recordings (\underline{Pitt}): 184 HC, 214 AD. Other pp information unreported. 
&
Clinical ast (i.e. medical, cognitive). \break
\textbf{Speech task:} narrative picture description (Cookie Theft). 
&
HC and AD. Based on cognitive ast (i.e. MMSE).
&
Acoustic analysis to extract paralingusitic fts directly from speech recordings, no ft selection, ML for group classification.
&
Detection performance:\break
68\% accuracy (baseline classification with simple algorithms for voice activity detection and speech rate).
\\

\citet{LUZ18.5}
&
Automatic detection of AD based on dialogical, content-free fts extracted from transcripts.
&
38 pps (\underline{CCC}, ibid.) 21 nonAD (12F), 17 AD (15F). Age over 65. Other demographics unreported.
&
\textbf{Speech task:} conversational interview about pp's chronic condition and their experience in healthcare.
&
HC (nonAD: patients with chronic conditions unrelated to AD) and AD. Based on clinical diagnosis.
&
Dialogue analysis for ft extraction from conversational transcripts, Markov chains for data representation, ML for group classification.
&
Detection performance:\break
86.5\% accuracy with vocalisations and speech rate.
\\

\citet{MartinezdeLizarduy2017}
&
Automatic detection of MCI and AD based on acoustic fts with a novel decision support system (ALZUMERIC).
&
\underline{SVF} ds: 62 HC (36F), 38 MCI (21F). Avg age: 56.73, 57.15.
\underline{PD} ds\footnote{Both are subsets from the Gipuzkoa-Alzheimer Project: http://www.cita-alzheimer.org/projects/gipuzkoa-alzheimer-project-basque-cohort}: 12 HC, 6 AD. 
SS ds (\underline{AZTIAHORE}): 20 HC (9F), 20 AD (12F).
&
\textbf{Speech tasks}: SVF (animals), picture description (PD), spontaneous speech (SS, see \underline{AZTIAHORE}).
&
SVF: HCand MCI. \break
PD: HC and AD.\break
SS: HC and AD.\break
Unreported criteria.
&
ALZUMERIC system: acoustic analysis for ft extraction from voice samples, automatic ft selection, ML for group classification.
&
Detection performance:
HC-MCI (SVF): 80\% accuracy.\break
HC-AD: 94\% (PD) and 95\% accuracy.
\\

\citet{Meilan2014}
&
Automatic detection of AD based on temporal and acoustic speech fts.
&
66 pps: 36 HC (80\%F)  30 AD (68\%FF). Avg age: 74.06, 78.66; educ: 7.30, 6.27; MMSE: 27.97, 18.07.
&
Cognitive ast.\break
\textbf{Speech task:} reading familiar sentences on screen.
&
HC and AD. Based on NINCDS-ADRDA and cognitive ast (i.e. GDS, MMSE).
&
Acoustic and temporal speech analysis for ft extraction, ML for group classification. 
&
Detection performance:\break
83.3\% accuracy with speech fts such as voice breaks.
\\

\citet{mirheidari2017toward}
&
Discrimination between ND and FMD\footnote{ND: neurodegenerative disorder (e.g. AD). FMD: functional memory disorder)}, based on doctor-patient conversational fts.
&
30 pps: \break
15 FMD (9F), 15 ND\footnote{Heterogeneous ND group: 8 AD, 3 AD+vD, 2 MCI and 2 FTD (frontotemporal dementia).} (8F). Avg age: 57.8, 63.73; MMSE: 28.87, 18.79.
&
Cognitive ast.\break
\textbf{Speech task:} neurology consultation (conversation).
&
FMD and ND. Based on \citet{schmidtke2008syndrome} (FMD), Petersen and NINCDS-ADRDA (ND)
&
Automatic conversation analysis for ft extraction, ML for ft selection and group classification.
&
Discrimination performance:\break
 97\% classification accuracy between FMD and ND with top-10 fts.
\\

\citet{mirheidari2018detecting}
&
Discrimination between different conditions based on linguistic fts from conversational speech.
&
\underline{IMDB}\footnote{DS details. IMBD: text entries on movies feedback. \underline{Pitt} (previously described). Hallam \cite{mirheidari2017toward}, IVA \cite{mirheidari2017avatar} and Seizure \cite{ekberg2015can}}.: 50000 entries.\break
\underline{Pitt}: 473 narratives.
\underline{Hallam}: 45 conversations. \break
\underline{IVA}: 18 conversations. \break
\underline{Seizure}: 241 conversations.
&
\textbf{Written task:} movie feedback (\underline{IMDB}).\break
\textbf{Speech task:} picture description (\underline{Pitt}), neurology consultation.
&
\underline{Pitt}: HC, AD.\break
\underline{Hallam}: FMD, ND, DPD.\footnote{DPD: depressive pseudo-dementia.}.\break
\underline{IVA}: FMD, ND, MCI.\break
Unreported criteria.
&
ASR and linguistic analysis (vector representation) for ft extraction and selection, ML for group classification.
&
Discrimination performance:\break
65.8\% (\underline{Hallam}), 70\% (\underline{IVA}).\break
Best binary: \break
93.7\% FMD-DPD (\underline{Hallam}),
    100\% FMD-ND (\underline{IVA}).\\

\citet{Mirheidari2019dementia}
&
Automatic detection of ND based on speech fts, comparing neurologist-led with virtual-agent-led interactions (\underline{IVA}).
&
\underline{HUM}, 30pps: 15 FMD (9F), 15ND\footnote{Heterogeneous ND group: 8 AD, 3 AD+vD, 2 MCI and 2 FTD (frontotemporal dementia).} (8F). Age: 57.8, 63.73; MMSE: 28.87, 18.79.\break
\underline{IVA}, 12 pps: 6 FMD (1F), 6 ND (3F). Avg age: 55.67, 65.83, ACE-R: 83.67, 59.57.
&
Cognitive ast.\break
\textbf{Speech task:} neurology consultation (\underline{HUM}) or avatar interaction (\underline{IVA}).
&
FMD and ND. Based on \citet{schmidtke2008syndrome} (FMD), Petersen and NINCDS-ADRDA (ND)
&
Acoustic, linguistic and conversational analysis of recordings, comparing IVA-patient with neurologist-patient interactions. ML for group classification.
&
Detection performance:\break
Neurologist-patient: 90.0\% accuracy.\break
IVA-patient: 90.9\% accuracy.
\\

\citet{Mirheidari2019computational}
&
Discrimination between HC, FMD, MCI and ND based on acoustic and linguistic fts from IVA-led speech.
&
61 pps: 14 HC (8F), 10 FMD (6F), 18 MCI (12F), 19 ND (7F). Avg age: 69.4, 56.4, 62.2, 69.8. 
&
Cognitive ast. \break
\textbf{Speech task:} SVF (animals) and 10 question conversations. Both led by an IVA.
&
HC, FMD, MCI and ND. Based Schmidtke, (FMD), Petersen (MCI) and NINCDS-ADRDA (ND) criteria.
&
Acoustic, linguistic and conversational analysis of SVF answers and IVA-patient interactions. ML for ft selection and group classification.
&
4-way discrimination performance:\break
62\% accuracy and ROC-AUC: 81.5\% with top 22 fts (48\% all).
\\

\citet{Mirzaei2018}
&
Discriminate between HC, MCI and AD based on acoustic fts from narrative speech.
&
48 pps: 16 HC, 16 MCI, 16 AD. Avg age: 72.7, 77.6, 77.9; MMSE: 28.6, 28.3, 22.4. 
&
Cognitive ast\break
\textbf{Speech task:} reading familiar sentences on screen.
&
HC, MCI, AD. Based on cognitive ast ((MMSE over 20 for inclusion).
&
Acoustic analysis for ft extraction, ML for ft selection (wrapper) and group classification. 
&
Discrimination performance:\break
62\% accuracy (three-way classification).
\\

\citet{Nasrolahzadeh2018}
&
Discriminate between HC and three stages of AD based on higher-order spectral analysis of speech data.
&
60 pps\footnote{30 AD participants distributed in three levels: First Stage (FS), Second Stage (SS) and Third Stage (TS).)}: 30 HC (15F), 6 FS (3F), 15 SS (6F), 9 TS (5F). Avg age: 75.6, 73.3, 70.6, 77.4; MMSE: 28.39, 27.5, 26.8, 23.8.
&
Clinical ast (i.e. cognitive, medical).\break
\textbf{Speech task:} prompted to talk about personal stories and feelings.
&
HC and AD, subdivided in FS, SS and TS. AD subgroups diagnosed with NINCDS-ADRDA criteria.
&
Acoustic spectral analysis for nonlinear feature extraction, ML for for ft selection and group classification.
&
Discrimination performance:
97.71\% accuracy with 4-way classifier based on higher-order spectral fts.
\\

\citet{orimaye2017predicting}
&
Automatic detection of AD based on linguistic fts extracted from narrative speech.
&
198 pps (\underline{Pitt}): 99 HC and 99 AD. Avg age: 65.26, 70.45. Other demographics unreported.
&
\textbf{Speech task:} narrative picture description (Cookie Theft). 
&
HC and ADUnreported criteria.
&
Linguistic analysis for ft extraction, statistical analysis for ft selection, ML for group classification.
&
Detection performance:
$AUC=0.93$ with 1000 top combined fts (syntacic, lexical, n-grams).
\\

\citet{PrudHommeaux2011}
&
Detection of MCI based on automatic alignment scores between transcribed recall tasks and source narratives.
&
124 pps: 52 HC, 72 MCI. \break
Demographics unreported.
&
Cognitive ast.\break
\textbf{Speech task:} immediate and delayed recall of the Anna Thomson story (LM-WMS-III\footnote{Logical Memory Test of the Weschler Memory Scale III \cite{wechsler1997wms}}).
&
HC (non-MCI) and MCI. Based on cognitive ast (i.e. CDR=0.5 for MCI diagnosis).
&
Linguistic analysis and text alignment for automatic scoring of recall tasks, ML for group classification.
&
Scoring performance:\break
$F_1=0.791$ (alignment based).\break
Detection performance:\break
$AUC=0.795\%$).\\

\citet{Prudhommeaux2015}
&
Detection of MCI based on automatic alignment scores between transcribed recall tasks and source narratives.
&
235 pps: 163 HC, 72 MCI. Avg age: 87.3, 88.7. Avg educ: 15.1, 14.9. Gender unreported.
&
Cognitive ast.\break
\textbf{Speech task:} immediate and delayed recall of the Anna Thomson story (LM-WMS-III).
&
HC (non-MCI) and MCI. Based on cognitive ast (i.e. CDR=0.5 for MCI diagnosis).
&
Linguistic analysis and graph-based text alignment for automatic scoring of recall tasks, ML for group classification.
&
Scoring performance:\break
$F_1=0.891$.\break
Detection performance:\break
$AUC=0.748$. \break
\underline{Pitt}:\footnote{This is an attempt from to authors to apply their model on unseen data. Only results based on their graph-based method are reported here.}: $AUC=0.704$
\\

\citet{Rentoumi2017}
&
Automatic detection of AD based on linguistic fts extracted from written narrative data.
&
60 pps: 30 HC (14F), 30 AD (17F). Avg age: 68.03, 66.48; educ: 13.93, 12; MMSE: 28.26, 22.68.
&
Cognitive ast.\break
\textbf{Written task:} narrative picture description (Cookie Theft).
&
HC (NC) and AD. Based on cognitive ast. ($MMSE_{AD}=10-25$).
&
Computational linguistic analysis for text ft extraction (morphosyntactic, lexical).\break
ML for group classification.
&
Detection performance:\break
80\% accuracy.\break
(88.5\% with synthetically enlarged ds).
\\

\citet{Roark2011}
&
Automatic detection of MCI based on scores and speech fts from recorded cognitive tests.
&
74 pps: 37 HC, 37 MCI. Avg age: 88.8, 89.8; educ: 15.1, 14.5; MMSE: 28.2, 26.4. Gender unreported.
&
Cognitive ast.\break
\textbf{Speech task:} immediate and delayed recall of the Anna Thomson story. 
&
HC and MCI. Based on cognitive ast (i.e. CDR=0.5 for MCI diagnosis).
&
Linguistic and acoustic analysis for ft extraction, statistical analysis for ft selection, ML for group classification.
&
Detection performance:
$AUC =0.861$ (test scores and automatically derived speech and language fts).
\\

\citet{Rochford2012}
&
Automatic detection of CI based on pause distribution fts from narrative speech.
&
187 pps: 150 HC, 37 CI. Avg age (all): 72.44; MMSE: 27.68, 114 females. educ unreported. 
&
Cognitive ast.\break
Speech task: ready aloud a passage from a children's story.
&
HC and CI. \break
Based on cognitive ast. ($MMSE_{HC}>=27$, $MMSE_{CI}<27$).
&
Linguistic and acoustic analysis for ft extraction, statistical analysis for ft selection, ML for group classification.
&
Detection performance:\break
68.66\% acc ($AUC=0.74$).\break
\\

\citet{Sadeghian2017}
&
Automatic detection of AD based on acoustic and linguistic fts from narrative speech, and with MMSE scores.
&
72 pps: 46 HC, 26 AD. Avg age: 71.43, 78.48; educ: 13.28, 13.81; MMSE: 28.70, 20.92.
&
Cognitive ast.\break
\textbf{Speech task:} narrative picture description (new picture\footnote{https://acoustics.org/wp-content/uploads/2015/10/Sadeghian-Figure1b.jpg}).
&
HCand AD. Based on medical diagnosis.
&
Customised ASR for speech transcription, acoustic and linguistic analysis ft extraction, ML for ft selection and group classification.
&
Detection performance (acc):\break
88.3\%: demogr.+acoustic.\break
91.7\%: ASR linguistic.\break
94.4\%: MMSE+manual ling
\\

\citet{Satt2013}
&
Discrimination between HC, MCI and AD based on acoustic fts (content-free) from voice recordings.
&
\underline{Dem@care}: 89 pps. \break 19 HC (15F), 43 MCI (31F), 27 AD (24F). Avg age: 67, 73, 72. 
&
\textbf{Speech task:} narrative picture description, sentence repetition (15), syllable repetition ("pa-ta-ka").
&
HC, MCI and AD.\break
Based on medical diagnosis.
&
Speech segmentation (VAD), acoustic analysis for ft extraction, statistical analyis for ft selection, ML for group classification.
&
Discrimination performance:
$EER_{HC-MCI+AD}=18\%$\break
$EER_{HC-MCI}=17\%$\break
$EER_{HC-AD}=15.5\%$
\\

\citet{Shinkawa2019}
&
Automatic detection of MCI based on single modality and multimodal behavioural data (gait and speech).
&
34 pps: 19 HC (12F), 15 MCI (8F). Avg age: 71.63, 74.87; MMSE: 28.42, 25.33.
&
Clinical ast.\break 
\textbf{Speech task:} narrative picture description (Cookie Theft).\break
Gait task: 5-meter walk.
&
HC and MCI. Based on Petersen criteria.
&
ASR for speech transcription, gait and linguistic analysis for ft extraction, statistic analysis and ML for ft selection and group classification.
&
Detection performance:\break
Multimodal: 82.4\% acc.\break
Single modality: 76.6\% acc each ($F_{1speech}=0.733$, $F_{1gait}=0.667$).
\\

\citet{Tanaka2017}
&
Automatic detection of CI based on audiovisual fts from dialogues with a computer avatar.
&
29 pps: 15 HC (4F), 14 CI\footnote{Heterogeneous CI group: 9 AD, 1 NPH (normal pressure hydrocephalus), 1 AD+NPH, 1 DBL (dementia with Lewy bodies)} (4F). Avg age: 74.1, 76.3; educ: 10.5, 14.1; MMSE: 27.5, 21.4.
&
Cognitive ast.\break
\textbf{Speech task:} 10-15 min interaction with an avatar (dialogue system).
&
HC and CI.\break
Based on medical diagnosis.
&
Acoustic, linguistic and image analysis for ft extraction, statistics for ft selection, ML for group classification. 
&
Detection performance:\break
83\% unweighted acc. \break
$AUC=0.93$.\break
$AUC_{ADonly}=0.89$.
\\

\citet{Thomas2005}
  &
 Discrimination of different stages of CI based on linguistic fts from interviews.
  &
 95 pps (\underline{ACADIE}\footnote{ACADIE study \cite{rockwood2004acadie}. Pps are divided by MMSE in either 2 (high, low) or four groups (normal, mild, moderate, severe)}): 85 high, 73 low; 35 normal, 50 mild, 53 moderate, 20 severe.
 &
 Cognitive ast.
 \textbf{Speech task:} two interviews about donepezil, 12 weeks apart.
 &
 Two or four groups. Based on MMSE scores (0-15, 16-20, 21-24 and 25-30).
  &
 Linguistic analysis for ft extraction (lexical, n-gram), statistics for ft selection, ML for group classification.
&
Discrimination performance:
95\% acc severe vs. normal.\break
69.6\% moderate vs. mild.\break
50\% 4-way classification.\\

\citet{Toth2018}
&
Automatic detection of MCI based on acoustic fts from narrative speech
&
84 pps: 36 HC (23F), 48 MCI (32F). Avg age: 64.13, 73.08; educ: 12.47, 11.82; MMSE: 29.17, 26.97; 
&
Cognitive ast.\break
\textbf{Speech task:} previous day, immediate/delayed recall of 2 short films.
&
HC and MCI. Based on cognitive ast (i.e. MMSE, CDT, ADAS-Cog)
&
Customised ASR and acoustic analysis for ft extraction, statistics for ft selection, ML for gorup classification.
&
Detection performance:\break
75\% acc automatic procedure ($F_1=0.788$, $AUC=o.676$) .
\\

\citet{Troger2017}
&
Automatic detection of AD based on acoustic fts from narrative speech.
&
\underline{Dem@Care}: 115 pps.\break 47 HC (40F), 68 AD (38F). Avg age: 72.4, 78.9. 
&
Cognitive ast.\break
\textbf{Speech task:} two life events, previous day, picture description.
&
HC and AD\footnote{Heterogeneous group: diagnosed with either AD or a form of mixed dementia (including AD).}\break
Based on medical diagnosis.
&
Acoustic signal processing for ft extraction, univariate ft selection, ML for group classification.
&
Detection performance:\break
89\% acc relying solely on vocal fts (ASR and content-free).
\\

\citet{Troger2018}
&
Discrimination between SCI, MCI and AD with a simulated telephone-based SVF test (feasibility study).
&
166 pps: 40 SCI (32F), 47 MCI (24F), 79 AD (40F). Avg age: 72.65, 76.59, 79; educ: 11.35, 10.81, 9.47; MMSE: 28.27, 26.02, 18.81; 
&
Clinical ast.\break
\textbf{Speech task:} recorded SVF answers (animals).
&
SCI, MCI and AD. Based on subjective reports (SCI), Petersen (MCI) and NINCDS-ADRDA (AD)
&
ASR, acoustic and linguistic analysis for ft extraction, ML for group classification.
&
ASR performance:\break
VFER\footnote{VFER: Verbal Fluency Error Rate}$=33.4\%$.\break
$AUC=0.855$
\\

\citet{Weiner2016a}
&
Automatic detection of AD based on acoustic fts from conversational speech in German.
&
\underline{ISLE}: 74 pps\footnote{subset from \underline{ILSE}: Interdisciplinary Longitudinal Study on Adult Development and Aging \cite{weiner2016towards}.}.\break 98 samples: 80HC, 13 AACD. 5AD. Age range: 70-74 years old.
&
Clinical ast.\break
\textbf{Speech task:} semi-standardized biographic interviews.
&
HC, AACD (ageing associated cognitive decline) and AD.
Based on medical diagnosis.
&
Acoustic analysis for ft extraction (focus on pause patterns), ML for group classification.
&
Detection performance:\break
85.7\% acc. $UAR=0.66$\break
$F_{1_{HC}}=0.92$,$F_{1_{AD}}=0.80$,$F_{1_{AACD}}=0.80$.
\\

\citet{Weiner2016}
&
Automatic prediction of the development of CI from conversational speech in German.
&
\underline{ISLE}: 51 pp. \break 35 HC, 16 CI\footnote{Heterogeneous group: AACD, MCD (mild cognitive disorder), AD, VAD (vascular dementia)} (developed within three visits). Age range: 61-77 (1st-3rd visit).
&
Clinical ast.\break
\textbf{Speech task:} semi-standardized biographic interviews.
&
\textit{No change} (HC) and \textit{Change} (CI). Based on whether they remained healthy or not.
&
VAD and acoustic analysis for ft extraction (focus on pause patterns), ML for group classification.
&
Prediction performance:\break
80.4\% acc (overall). 
$Rc_{no-change}=0.91$, $Rc_{change}=0.56$
\\

\citet{Yu2018}
&
Automatic detection of CI based on speech fts collected through remote assessments.
&
\underline{ADCS\footnote{ADCS: Alzheimer's Disease Cooperative Study. 4-year longitudinal data collection for home-based assessment.}}: 167 pps. \break Pre-processed 180 samples: 160 HC, 20 CI.
&
Clinical ast.\break
\textbf{Speech task:} SVF and EBi/EBd\footnote{East Boston Immediate/Delayed: summarise a story immediately/delayed after listening to it).} delivered by telephone system.
&
HC and CI. Based on longitudinal medical diagnosis.
&
Acoustic analysis for ft extraction (articulatory, phonemic), ML for ft selection and group classification.
&
Detection performance:\break
Speech+scores: AUC$=0.77$
Speech only: AUC$=0.74$ \break
Scores only: AUC$=0.54$
\\
\hline
\end{longtable}

\end{landscape}

\subsection{Data details table}
This table accounts for details of the datasets, as well as specific
subsets, used in the reviewed studies.  It is structured as follows:
 \begin{itemize}
 \item \textbf{Data set size}: number of participants or samples,
   including details on number of words, or number of hours recorded,
   when available.
 \item \textbf{Data type}, with two distinctions: a) writings, audio
   recordings and/or transcripts (abbreviated as per Table~\ref{tab:textualabbrev}); b)
   monologues or dialogues. Monologues, in turn, are divided into
   spontaneous, narratives and answers to cognitive tests (most
   frequently fluency task), whilst dialogues are subdivided into
   three groups: structured, semi-structured and conversational. When
   available, information about transcription (i.e. software used, 
   manual vs. automatic) is included.
 \item \textbf{Other modalities}: such as video, cognitive scores or
   motor measurements, when applicable ("NA" is written otherwise).
 \item \textbf{Data annotation}: group labels available in the data,
   corresponding with what was described in the comparison groups
   column of the SPICMO table. It includes groups' \textit{n},
   i.e. group size, as well as groups' \textit{m}, i.e. number of
   speech/test samples per group, as sometimes these two figures
   differ (e.g. in longitudinal studies).
 \item \textbf{Data balance}: whether the dataset or subset used in
   the study is balanced in terms of age, gender and education. It
   accounts for dataset balance, within class balance and between
   class balance when applicable (see ``keys to table
   interpretability'', above, for acronyms). If a feature is not
   reported in the table, this is because it was not reported in the
   article.
 \item \textbf{Data availability}: whether the data used in the study
   is available to the wider research community.
 \item \textbf{Language}: language in which the dataset was collected,
   including country of origin, since many languages are spoken in
   more than one country.
 \end{itemize}
    
 Names of particular datasets are underlined (e.g. \underline{Pitt})
 The second table aims to provide the community with benchmark
 information about current databases and their availability, in order
 to highlight recurrent gaps that future research projects should
 target when designing their data collection procedures.

\newpage
 \begin{landscape}
   \setlength{\footskip}{3cm}
\scriptsize
\newcolumntype{R}[1]{>{\raggedright\arraybackslash}p{#1}}
\setlength\LTleft{-2.2cm}
\begin{longtable}{R{1.6cm}R{2.8cm}R{3.3cm}R{2.3cm}R{2.3cm}R{2.8cm}R{2.8cm}R{1.8cm}} 
  \caption{\label{tab:data} Detailed Data information} \\\toprule
  \textbf{Study} & \textbf{Data set/Subset size} & \textbf{Data type } &
  \textbf{Other modalities} & \textbf{Data annotation}  
  & \textbf{Data balance} & \textbf{Data availability} & \textbf{Language}\\\midrule
\endfirsthead
  \caption{Detailed Data information (ctd.)} \\\toprule
  \textbf{Study} & \textbf{Data set/Subset size} & \textbf{Data type } &
  \textbf{Other modalities} & \textbf{Data annotation}  
  & \textbf{Data balance} & \textbf{Data availability} & \textbf{Language}\\\midrule
\endhead

\citet{Beltrami2016}
&
39 pps. 
&
Narrative monologues.\break
Rec. and manual tr. (Transcriber\footnote{http://trans.sourceforge.net}).
&
Cognitive scores: MMSE, MoCA, GPCog, CDT, VF.
&
HC (CON) / MCI: \break
$n=20/19$.
&
CB.\break
Text reports balanced demogr., but no figures.
&
Unreported.
&
Italian (Italy).\\

\citet{BenAmmar2018}
&
\underline{Pitt:} $m=484$.\break
No. of pps unreported.
&
Narrative monologues.\break
Rec. and manual tr. (CHAT\footnote{CHAT protocol: Codes for the Human Analysis of Transcripts \cite{brian2000childes}.}).
&
Cognitive scores: MMSE.
&
HC / AD (Dementia):\break
$m=242/242$.
&
CB.\break
Demogr. unreported.
&
\underline{Pitt} avail. (DementiaBank).\break
Enhanced unreported
&
English (US).
\\

\citet{Bertola2014}
&
100 pps.
&
Monologues: fluency task.\break
Rec. (unclear).
&
Cognitive scores: MMSE, Katz, Lawton, SVF.
&
HC (NC), aMCI, a+mdMCI\footnote{aMCI: amnesic single-domain; a+mdMCI: amnesic multiple-domain. Class-balance depends on whether they are considered 1 or 2 groups.}, AD: $n=25$
&
no-CB/CB (ibid.).\break
no-WCGB: HC.\break
BCB: A, G, E.
&
Unreported.
&
Portuguese (Brazil).\\

\citet{Chien2018}
&
60 pps: 30 HC \textit{ad-hoc}, 30 AD (Mandarin\_Lu). 
3 tasks each: $m=150$.
&
Narrative monologues and fluency task.\break
Rec.
&
Cognitive scores: SVF.
&
HC (CH) / AD: \break
$n=30/30$;\break 
$m=75/75$.
&
CB.\break
Demogr. unreported.
&
Mandarin\_Lu avail. (DementiaBank).\break
HC unreported.
&
Chinese, Taiwanese (Taiwan).\\

\citet{Clark2016}
&
158 pps.
&
Monologues: fluency task.\break
Manual tr. (text file).
&
Cognitive scores: CDR, MMSE, SVF.\break
NI meas.: MRI.
&
HC (CN)/MCI-non\footnote{MCI-non (non converters) and MCI-con (converters) refer to whether MCI pps converted to AD or not over a 4-year follow-up.}/MCI-con: $n=51/83/24$.
&
No-CB, no-GB.\break
no-WCGB.\break
BCB: no-A, no-G, E.
&
Unreported.
&
English (US).
\\
\hline
\end{longtable}

 \end{landscape}

\newpage

\subsection{Methodology table}
This table summarises the features and methods employed in the
reviewed studies. It is structured as follows:

\begin{itemize}
\item \textbf{Pre-processing}: where available, this column describes the
  procedures undertaken on text and audio data as preparation steps
  for subsequent analysis. before. For text, this includes
  transcription (manual or ASR), tokenisation, removal of unanalysable
  events and \textit{stopwords}, and so on. For audio, this includes
  background noise removal, normalisation, speaker diarisation.
\item \textbf{Feature generation}: whether the features were generated from raw data through text analysis and/or through acoustic analysis, followed by more specific subcategories as per the taxonomy described in Table~\ref{tab:FeatureTaxonomy}. When reported, this column also includes the paper's approach to reduce the extracted feature set, essentially either selection or extraction. On the one hand, \textit{'filtering' selection} uses extrinsic criteria, such as information gain or, commonly, \textit{p}-values (i.e. whether the differences between the experimental groups, e.g. AD and HC, for a particular feature are statistically significant or not); whereas \textit{'wrapping' selection} uses a cross-validation model that searches through the power set of features. On the other hand, \textit{feature extraction} entails creating a new reduced feature set by combining or transforming the original one with method such as PCA, LSA, clustering or ADR. 
\item \textbf{ML task/method}: supervised vs. unsupervised
  learning. Task: clustering, classification, regression. Method:
  clustering algorithm, classifiers and regression method as per Table~\ref{tab:abbrMethods+Metrics}. This column also
  includes
  information on the number of classes that the classifier outputs.
\item \textbf{Evaluation technique}: describes four points, when
  available. First, the baseline against which the study results are
  compared (i.e. random guess, neuropsychological scores, different
  feature sets). Second, the performance metrics reported by the
  authors (i.e. \textit{acc, F1, pc, rc, ss, sp, AUC, EER}, see Table~\ref{tab:abbrMethods+Metrics}). This will
  include information about different ASR precision measures, such as
  WER, where applicable. Third, the cross-validation technique used. Fourth, whether a test
  set held out, unused for model training, and its size.
\item \textbf{Results}: numerical results of the selected performance
  metrics for the baseline and for the fitted model/s. When multiple
  metrics are reported, only summary metrics such as \textit{EER, acc,
    F1} and \textit{AUC} are included in this column.
\end{itemize}

\newpage
\begin{landscape}
\scriptsize
\enlargethispage{3.2\baselineskip}
\newcolumntype{R}[1]{>{\arraybackslash}p{#1}}
\setlength\LTleft{-2.5cm}
\setlength\LTpre{0pt}
\setlength\LTpost{0pt}
\begin{longtable}{R{1.7cm}R{4.2cm}R{5cm}R{3.5cm}R{3.5cm}R{3.5cm}} 
\caption{\label{tab:methods}Methodology} \\\toprule
\textbf{Study} & \textbf{Pre-processing} & \textbf{Feature generation} & \textbf{ML task/method} & \textbf{Evaluation technique} & \textbf{Results}
\\\midrule
\endfirsthead
\caption{Methodology(ctd.)} \\\toprule
\textbf{Study} & \textbf{Pre-processing} & \textbf{Feature generation} & \textbf{ML task/method} & \textbf{Evaluation technique} & \textbf{Results}
\\\midrule
\endhead

\citet{Beltrami2016}
&
Processing unit: utt \newline 
\textbf{Text:} manual tr. 
Paralinguistic annotation.\newline
\textbf{Audio:} VAD (ssvad\footnote{VAD proposed by \cite{mak2014study}}) and Kaldi\footnote{http://kaldi.sourceforge.net/about.html}-ASR forced alignment
&
Filtering (selection): \textit{p-values}.\newline
\textbf{Text-based:} lex diversity, PoS, lex density, syntactical (dependency).\newline
\textbf{Acoustic:} prosodic (temporal, $F_0$, energy), spectral.
&
Supervised learning.\newline
\textbf{Classification}: binary (HC-MCI) with \textit{k}-NN ($k=3$), LR and NN.
&
B/L: unreported.\newline
Metrics: \textit{acc, pc, rc, $F1$}.\newline
CV: unreported.\newline
Hold-out set: 80/20\%.
&
LR and NN performed best on "Picture" task:
$acc=76.9\%$, $pc=0.727$, $rc=0.842$ and $F1=0.781$.
\\


\citet{BenAmmar2018}
&
\textbf{Text:} ASR tr.\newline
\textbf{Audio:} removal of background noise and non-analysable\footnote{Non-analysable events in this context refers to breaks, overlapping speech, coughing, laughter, short hard noises and the like.} events.
&
Filtering: IG; Wrapping: \textit{k}-NN, SVM.\newline
\textbf{Text-based:} lex diversity, lex density, syntactical (constituency), pragmatics (\textit{UoL}).
&
Supervised learning.\newline
\textbf{Classification}: binary (HC-AD) with NN, SVM and DT.
&
B/L: no ft set reduction.\newline
Metrics: \textit{acc}.\newline
CV: unreported.\newline
Hold-out set: unreported.
&
Best performance: $acc=79\%$ SVM.\newline
Best ft set: \textit{k}-NN ($acc=69\%$ NN, $71\%$ DT).
\\
\hline

\end{longtable}

\end{landscape}


\subsection*{Clinical applicability}
This table summarises our assessment of the potential implications and
applications of findings of each reviewed paper as regards research
and clinical use. The table is structured as follows: 

\begin{itemize}
\item \textbf{Research implications}: 
\begin{itemize}
\item \underline{Research Novelty}: whether at the time of publication the study described a new dataset, proposed a new set of features, implemented a new method or applied an existing one for a different task;
  \item \underline{Study Replicability}: \textit{low}, \textit{partial} or \textit{full}, depending on how well the procedure is described
  and whether data or data identifiers are available). \textit{Low}
    refers to cases where both data is unavailable and method
    description is incomplete or unsatisfactory; \textit{partial} to cases
    where either is the case, and \textit{full} when both data and methods
    are available and satisfactorily described.
    \item \underline{Results generalisability}: \textit{low}, \textit{moderate} and \textit{high}, depending on whether the analysis is specific to the task, and/or there have been any extrinsic validation procedures and/or robust evaluation techniques are in place (i.e. train-test, CV, baseline). \textit{Low} refers to cases where the analysis is indeed specific to the task, and therefore difficult to apply to other tasks (e.g. when relying heavily on content features). In \textit{low} generalisability studies there are no extrinsic validation procedures (e.g. pilot in clinical settings) and the evaluation techniques are insufficient (e.g. CV is in place, but no train-test and/or appropriate baseline comparisons). The improvement of one of these features would bring the study up to \textit{moderate}, and further improvements would make its generalisability \textit{high}. Given the state of the field, no study is 100\% generalisable, hence why we have used this terminology instead of the same we used for replicability. For generalisability to be \textit{high}, most conditions need to be met except for the extrinsic validation, since it is still very uncommon in the field that studies are carried within a clinical setting.

  \end{itemize}
    
\item \textbf{Clinical potential}: external validation is outlined if present. That is, whether
  the procedure has actually been attempted in real life (\textit{yes}); or is,
  at least, embedded in a device, or the experimental design envisions realistic clinical testing at some stage (\textit{in-design}). This column also includes
  potential applications (i.e. early screening for new cases of SCI or
  similar, monitoring disease progression or supporting diagnosis of
  MCI and AD), potential outcomes for global health (i.e. language of
  study) and potential for the methodology to be remotely applicable
  (no, suggested potential, yes when tried or purposefully designed
  with that in mind).
    
\item \textbf{Risk of bias}: Feature balance (\textit{no/partial/yes}),
  suitable metrics (\textit{yes/no}, i.e. whether metrics other than overall
  accuracy are reported when data are class-imbalanced),
  contextualized results (\textit{yes/no}, i.e. whether an appropriate baseline
  is provided in order to put results into perspective), overfitting
  (\textit{yes/no}, i.e. whether cross-validation and/or hold-out set
procedures are implemented). With regards to sample size, we specify three ranges that ranges: $ds\leq50$, $ds\leq100$ and $ds>100$. 
  
\item \textbf{Strengths/Limitations}: several characteristics are
  listed with a yes/no answer, "yes" indicating strength and "no"
  indicating limitation. These characteristics are: 
  \begin{itemize}
    \item spontaneous speech: speech data is naturally generated, generated in response to an open-answer question or a narrative task, or generated in response to a scripted cognitive task (i.e. verbal fluency or counting).  Speech is considered spontaneous when it is natural and when its prompted by open-answered or narrative tasks. That is, for example, the Cookie Theft picture description would be spontaneous (although not natural), whereas reading sentences from a screen saying as many animals as possible within 60 seconds is not spontaneous (nor natural).
    \item conversational speech: whether the study includes dialogue data or only monologue.

    \item automation: the only characteristic that observes a 'middle' stage. Method automation can be labeled as {\em no}, when the only automated procedure is the ML task; {\em partial}, when aspects of the procedure other than the ML task, such as feature set reduction, are also automated;
or {\em total}, when everything is automated including
  preprocessing (e.g. ASR is used for transcription).

\item content-independence: whether the model for feature generation relies heavily on content features of the data (e.g. lexical or high level \textit{n}-gram are often closely related to the way in which spoken language was prompted).
  \item Transcription-free: text analysis usually requires transcripts. Whether manual or ASR, transcribing procedures entail many restrictions. Manual transcription is time-consuming, whereas ASR transcription have limited performance on impaired speech, and they need to be trained to a specific language, therefore adding an extra step to the method. 
    \end{itemize}
\end{itemize}

\newpage
\begin{landscape}
  \setlength{\footskip}{3cm}
\scriptsize
\newcolumntype{R}[1]{>{\raggedright\arraybackslash}p{#1}}
\setlength\LTleft{-2.5cm}
\renewcommand*{\arraystretch}{1.4}
\begin{longtable}{R{1.5cm}R{5.5cm}R{6cm}R{4.8cm}R{4.5cm}}
  \caption{\label{tab:clinical}Clinical applicability} \\\toprule
  \textbf{Study} & \textbf{Research implications} & \textbf{Clinical
    potential}& \textbf{Risk of bias} & \textbf{Strengths/Limitations}
  \\\midrule
   \endfirsthead
   \caption{Clinical applicability (ctd)} \\\toprule 
\textbf{Study} & \textbf{Research implications} & \textbf{Clinical potential}& \textbf{Risk of bias} & \textbf{Strengths/Limitations} \\\midrule 
\endhead

\citet{Beltrami2016}
&
\textbf{Novelty:} preliminary results of new project (OPLON).\break
\textbf{Replicability:} partial.\break
\textbf{Generalisability:} low (task-specific model)
&
\textbf{External validation:} no.\break
\textbf{Potential application:} diagnosis support.\break
\textbf{Global Health:} Italian sentences.\break
\textbf{Remote application:} no.
&
\textbf{Feature balance:} yes.\break
\textbf{Suitable metrics:} yes (\textit{acc}).\break
\textbf{Contextualised results:} no.\break
\textbf{Overfitting:} hold-out set, no CV.\break
\textbf{Sample size:} $ds\leq50$ ($n=39$)
&
\textbf{Spontaneous speech:} yes.\break
\textbf{Conversational speech:} no.\break
\textbf{Automation:} partial (manual tr).\break
\textbf{Content-independence:} no.\break
\textbf{Transcription-free:} no.
\\

\citet{BenAmmar2018}
&
\textbf{Novelty:} speech samples only. Compare three ft selection processes.\break
\textbf{Replicability:} partial (unreported \textit{n} ).\break
\textbf{Generalisability:} low (task-specific model)
&
\textbf{External validation:} no.\break
\textbf{Potential application:} diagnosis support.\break
\textbf{Global Health:} US English sentences (\textit{Pitt}).\break
\textbf{Remote application:} no.
&
\textbf{Feature balance:} yes.\break
\textbf{Suitable metrics:} yes (\textit{acc}).\break
\textbf{Contextualised results:} yes.\break
\textbf{Overfitting:} no hold-out set, no CV.\break
\textbf{Sample size:} $ds>100$ ($m=484$)
&
\textbf{Spontaneous speech:} yes.\break
\textbf{Conversational speech:} no.\break
\textbf{Automation:} partial (manual tr).\break
\textbf{Content-independence:} no.\break
\textbf{Transcription-free:} no.
\\
\hline

\end{longtable}

\end{landscape}

\end{document}